\documentclass[11pt]{article}
\usepackage{authblk}
\usepackage{cite}
\usepackage{everysel}
\EverySelectfont{%
	\fontdimen2\font=0.4em% interword space
	\fontdimen3\font=0.2em% interword stretch
	\fontdimen4\font=0.1em% interword shrink
	\fontdimen7\font=0.1em% extra space
	\hyphenchar\font=`\-% to allow hyphenation
}
\makeatother
\usepackage{float}
\usepackage[english]{babel}
\floatstyle{plaintop}
\restylefloat{table}
\usepackage{graphicx}
\usepackage{booktabs}
\usepackage[linesnumbered,ruled]{algorithm2e}
\usepackage[caption = false]{subfig}
\usepackage{latexsym}
\usepackage[subfigure]{tocloft} 
\usepackage{multirow}
\usepackage{algorithmic}
\usepackage{verbatim}
\usepackage{moreverb}
\providecommand{\keywords}[1]
{
	\small	
	\textbf{\textit{Keywords---}} #1
}
\usepackage{hyperref}
\usepackage{listings}
\usepackage{float}
\usepackage{graphics}
\usepackage{algorithm2e}
\usepackage{paralist}
\usepackage{amssymb}
\usepackage{listings}
\usepackage{setspace}
\usepackage[T1]{fontenc}
\usepackage{textcomp}
\usepackage[utf8]{inputenc}
\usepackage{csquotes}
\usepackage{newunicodechar} 
\usepackage{lstautogobble}
\usepackage{times}
\usepackage{latexsym}
\usepackage{lipsum}
\usepackage{calligra}
\usepackage{verbatim}
\usepackage{enumerate}
\usepackage{pgfplots}
\newtheorem{challenge}{Challenge}
\newtheorem{finding}{Finding}
\usepackage{color}
\usepackage{tabulary}
\usepackage{tabularx}
\usepackage{siunitx}
\usepackage{csquotes}
\restylefloat{table}
\newtheorem{definition}{Definition}

\title{HCqa: Hybrid and Complex Question Answering on  Textual Corpus and Knowledge Graph}
\author[1]{Somayeh Asadifar}
\author[2]{Mohsen Kahani}
\author[3]{Saeedeh Shekarpour}
\affil[1,2]{Ferdowsi University, Mashhad, Iran}
\affil[3]{University of Dayton, Dayton, Ohio}
\date{}                     %% if you don't need date to appear

\begin{document}
	\maketitle
	\begin{abstract}
		Question Answering (QA) systems ease access to the vast amount of ever-growing data.
		In recent years, the research community has provided ad hoc solutions to the essential QA tasks, including named entity recognition, disambiguation, relation extraction, and query building. 
		However, the existing solutions are limited to simple and short questions whereas addressing complex questions which contain several sub-questions is neglected.
		The complex questions challenge NLP tasks such as language understanding, disambiguation,  relation extraction, and answer exploitation. Especially exploiting answer is further challenged if it requires a hybrid approach meaning integrating data from both corpus and knowledge graph.
		In this paper, we contribute to introducing an approach called HCqa which deals with complex questions requiring federation from a hybrid of corpus and knowledge graph. We contribute in developing (i) a decomposition mechanism to break down the input question into atomic sub-questions, (ii) a novel and comprehensive schema, first of its kind, for extracting relations and generating triples, and (iii) an approach for exploiting and federating the answers of sub-questions. 
		 Our experimental study exceeds the state-of-the-art in the fundamental tasks, such as relation extraction, as well as the answer set aggregation task which result in a good performance for answer extraction task.

	\end{abstract}
	
	
	%\todo[inline]{see this sentence: it has the accuracy near to state-of-art over ComplexQuestion benchmark
	%	it should be consistent to the previous report, how much accuracy? 
	%		and by accuray you mean F-score?
	%	}	
	
	%	\todo[inline]{also by query inventory you mean which gold standard, since you named the others, better to name all not some of them
	%	}	
	
	\section{Introduction}
	
	The Web of Data contains a wealth of knowledge (it currently includes more than 149 billion triples from 9,960 Knowledge Graphs (KG)\footnote{observed on 17 September 2018 at \url{http://lodstats.aksw.org/}}) belonging to a large number of various domains. 
	Although the Web of Data is a precious source for exploiting and retrieving informational need, it is still limited to encyclopedic information (e.g., Wikipedia) and then it is not adequately competent for up-to-date, real-time and fresh information.  
	However, a significant portion of the Web content consists of textual data from social network feeds, blogs, news, logs, etc.
	Thus, leveraging a hybrid of interlinked data sets from the Web of Data and textual content on Web might address more sufficiently for informational purposes.
	
	Question Answering (QA) interfaces enable end-users to interact via natural language regardless of concerns related to the structure of data (a single KG or multiple interlinked KGs or textual content), and background schema.
	However, the transformation of natural language into a formal representation is a challenge of high importance. It can be more challenging in case of complex questions which are typically longer, more ambiguous and require exploiting answers from a hybrid of KGs and textual content.
	A QA system federating knowledge from various heterogeneous sources is called a \textbf{hybrid} QA system \cite{bast2007ester}. 
	Despite the growth of semantics-enhanced and structure-empowered data, the state-of-the-art research community in QA systems still deals with short and simple queries, and it seems there is a long way to address long and complex questions \cite{singh2018frankenstein,sina1,lopez2009cross,unger2014question}.
	
	Particularly in specific domains such as the bio-medical domain, where the main body of knowledge resides in text. Thus, QA interfaces have to aggregate information from both KG and text to be qualified in real scenarios.
	The lack of a hybrid approach is a contributing factor in the majority of QA failures.
	To support the importance of hybrid QA systems, we further point out QALD \cite{unger2014question} challenge, held annually from 2011. It includes a specific track for hybrid QA systems, where the associated questions are relatively long and complicated and require integrating heterogeneous sources to exploit the final answer.
	
	In this paper, we contribute to developing a hybrid QA approach dealing with complex questions (i.e., long and complicated). Our approach is called \textbf{HCqa} which stands for ``Hybrid and Complex question answering''.
	To develop HCqa, this paper contributes to the following directions:

	\begin{itemize}
		%\item Investigating human thought process in exploiting answer of complex questions.
		\item Developing an approach for decomposing complex questions into atomic sub-questions. 
		\item A generic and comprehensive schema for relation extraction.
		\item Developing a methodology for exploiting answer from a hybrid manner.
		\item A detailed empirical study for measuring the effectiveness of our approach.
		
	\end{itemize}
	
	The rest of this paper is organized as follows. We review the state-of-the-art in Section 2.  The necessary preliminaries were introduced in Section 3. Section 4 presents a schema for relation extraction. Our approach is proposed in Section 5. Sections 6 shows the empirical results, and finally, we close with the concluding remarks and future work.
	\section{Related works}
	\label{sec:Related works}
	The literature related to QA, particularly for complex questions and hybrid approaches is spread in areas, such as information extraction, knowledge representation, and Natural Language Processing (NLP). In the following, we provide a brief overview of the state-of-the-art of these areas.
	
	\paragraph{\textbf{Information Extraction and NLP.}}
	Relation extraction is a critical task in information extraction, NLP and QA. It is concerned with recognizing relations between entities or between an entity and its associated attributes. This task is challenging because relations are mainly hidden, implicit or ambiguous. Literature in this area has two general directions, using a knowledge graph (KG) for relation extraction or not.\\ 
	Between approaches with the first direction, which were customized for question answering over KGs, some works rely on hand-crafted template or rules\cite{fader2013paraphrase}. Other works generate templates, automatically\cite{abujabal2017automated}. Recently, \cite{hu2018state} used a state transition framework to utilizing neural networks to answer complex question, searching for the sub-graph matching for the question.  In all of these approaches mapping to observed SPARQL queries, entities, predicates or sub-graphs in KG is a mandatory task. While in this paper, our key assumption is, the given complex question required textual corpus and knowledge graph federation to exploit the final answer; thus utilizing KG to generate relations is incorrect or insufficient for answer extraction. So our approach for relation extraction uses the way used by the second direction which is relation extraction over free text. \\In the second direction for relation extraction, the literature had focused on specific domains \cite{bunescu2005shortest,abacha2015means} using machine learning approaches, previously; thus the solutions were not easily applicable to other domains such as news and social media.
	Later approaches, such as Open Information Extraction (OIE), however, has eliminated some limitations \cite{yates2007textrunner,wu2010open,DelCorro2013,schmitz2012open}.
	OIE approaches are divided into two types of implementations (i) OIE, based on features like part of speech (POS) tags or shallow labeling of relations and their arguments, e.g., TextRunner \cite{yates2007textrunner} and WOE \cite{wu2010open}. Typically these implementations yield in uninformative and incoherent relations. 
	(ii) OIE implementations relying on deeper syntactic analysis e.g. Reverb \cite{fader2011identifying}, OLLIE \cite{schmitz2012open} and ClausIE \cite{DelCorro2013}. However, they are still unable to extract complex relations that even cannot be recognized by a dependency parser, e.g., LS3RyIE \cite{vo2017self} is an extension to ClausIE, where it modifies the structure of the dependency tree, generates patterns for relation extraction and then uses bootstrapping to learn more relations.
	However, a persisting deficiency is the length of arguments. \\NestIE \cite{bhutani2016nested} which is similar to OLLIE and WOE in training a dependency parser tree, addresses this deficiency using the concept of nested relations. NestIE first produces 13 patterns as seeds and then extracts paraphrases patterns using a dependency tree. Therefore, it could not address some sorts of complex relations, i.e., implicit relations, preposition-based relations, the comparative or superlative relations that their references are adjective phrases, question-based relations (How many questions), relations deduced from other simple relations (indirect verbal relation) and transforming n-ary relations into binary relations.\\
	Our strategy for relation extraction relies on a comprehensive linguistic background regarding the grammar of the English language, similar to the approaches presented in ClausIE \cite{DelCorro2013} and  LS3RyIE \cite{vo2017self}. ClausIE only extracts verbal and appositive relations from free text. Thuan et al. \cite{vo2017self} extends ClausIE with considering prepositional relations and pronoun relations without reference resolution.
	Our approach differs from the previous works since it relies on extracting triple patterns for different main elements of the grammar of English language and their subsets discussed in \cite{zandvoort2001handbook}, which are \texttt{Verb}, \texttt{Noun}, \texttt{Pronoun}, \texttt{Adjective} and \texttt{Adverb}. 
	In general, our approach extracts \emph{verbal}, \emph{Possesive Adjective+Whose}, \emph{Genitive  \& Preposition} and \emph{Appositive} relations related to questions i.e., \emph{Verbal} extended with \emph{Indirect Verbal}, the \emph{Preposition-based verbal}, \texttt{How many} and \texttt{How}+adjective relations. In addition, \emph{Noun phrase} and \emph{Comparative or Superlative} relations are extracted.
	With respect to \emph{Possesive Adjective+Whose} relations, we consider reference resolution. Also, in our approach, dependency parser \cite{Schuster2016} and Named Entity Recognizer (NER) \footnote{https://tagme.d4science.org/tagme/}$^{,}$\footnote{http://dbpedia-spotlight.github.io/demo/.} are employed.
	\paragraph{\textbf{QA approaches.}}
	QA community categorizes questions into simple and complex. Simple questions are factoid questions requiring the exploitation of the answer from a single information source without any particular constraint.
	In contrary, complex questions often need additional constraints for spotting answer. \cite {Kolomiyets2011} presents a brief overview of complex questions, such as list, descriptive, opinion, casual, and procedural. Also, \cite{oh2011compositional} defines complex questions as composite questions requiring federating information from various heterogeneous sources. 
	In general, researchers treat complex questions in two ways: (i) decomposition approaches \cite{hickl2004experiments,harabagiu2005employing} , (ii) segmenting approaches \cite{bast2012broccoli,saquete2004splitting,lopez2009cross,park2015isoft,usbeck2015hawk,oh2011compositional}. \\
	Broccoli \cite{bast2012broccoli} proposes a semi-automatic approach, which constructs the formal representation of composite questions by interacting with the user.
	The other approaches, which are fully automatic, rely on either shallow or deep linguistic analysis to segment the given composite question. 
	For instance, Saquete segments the composite question using a few pre-defined temporal signals and rules \cite{saquete2004splitting} or PowerAqua \cite{lopez2009cross,lopez2009merging} employs GATE \cite{cunningham2002framework} for tokenization, part of speech tagging and verb detection. 
	This information is used to generate relation triples. 
	SINA \cite{sina1, sina2} relies on a hidden Markov model to segment the given query concerning the background knowledge.
	ISOFT \cite{park2015isoft} recognizes a sequence of preposition or predicate phrases as sub-questions. 
	This approach is common for composite questions having sub-questions exposed in the predicate or prepositional phrases.
	Usbeck et al. \cite{usbeck2015hawk} and Xu et al. \cite{feng2016hybrid} extract sub-questions using a dependency parser; while no ordering for sub-question execution is determined and only a limited of number of patterns were used to sub-question generation. 
	Using lambda calculus is another approach to handle the composite question, e.g., \cite{frost2014denotational,frost2014event} considers each word as a function, and the semantics of the question is constructed using denotations and structural parsing results.\\  
	The best of the breed QA components are integrated and compared in the Frankenstein project \cite{Frankenstein1, Frankenstein2}. The Frankenstein framework was developed to allow the dynamic composition of QA pipelines based on the input question. It provides a full range of reusable components as independent modules of Frankenstein, populating the ecosystem leading to the option of creating many different components and QA systems. This project revealed the research gaps in the state-of-the-art tools for QA which our proposed research seeks to fill. 
	\section{Preliminaries}
	\label{sec:problemStatement}
	In this section, we present the necessary preliminaries and state the problem targeted in this work.
	Throughout the paper, we rely on the following four example questions which are taken from the hybrid task of QALD-6 benchmark\footnote{\url{https://qald.sebastianwalter.org/index.php?x=home&q=6}} \cite{unger20166th}. These example questions are used to illustrate our general approach, but different example questions and sentences used to explain the details of our approach.  
	\\
	Q1:\texttt{ Who was vice president under the president who approved the use of atomic weapons against Japan during World War II?}\\
	Q2:\texttt{ How many children does the actor who plays Dan White in Milk have?}\\
	Q3: \texttt{How many Golden Globes awards did the daughter of Henry Fonda win?} \\
	Q4: \texttt{Which recipient of Victoria Cross fought in the Battle of Arnhem?} \\
	
\begin{figure}[!h]
		\centering
		\includegraphics[width=\textwidth]{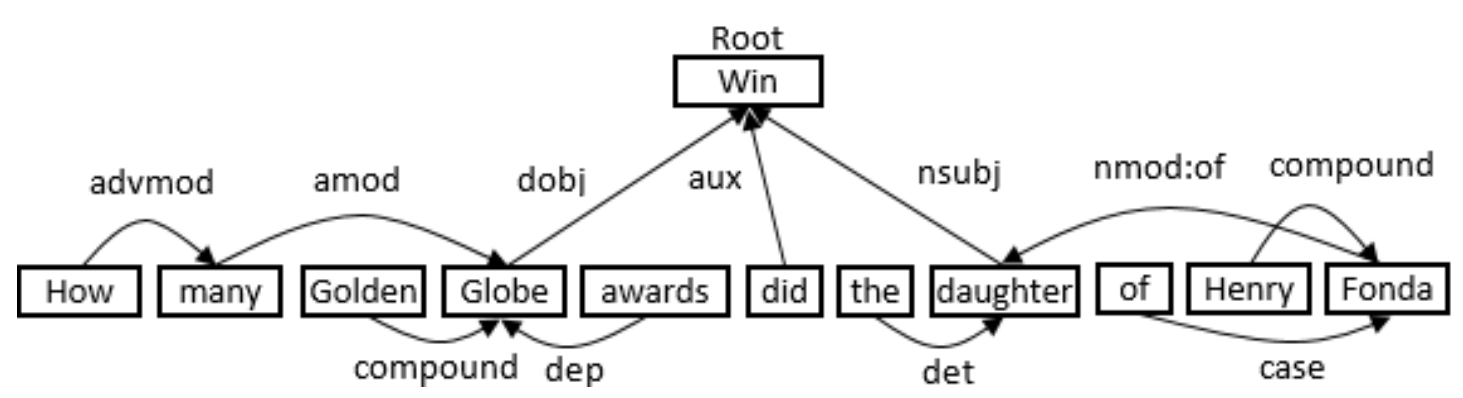}
		\caption{The dependency tree of Q3.}
		\label{fig:DP}
	\end{figure}

\begin{figure}[ht]
		\centering
		\includegraphics[scale=0.9]{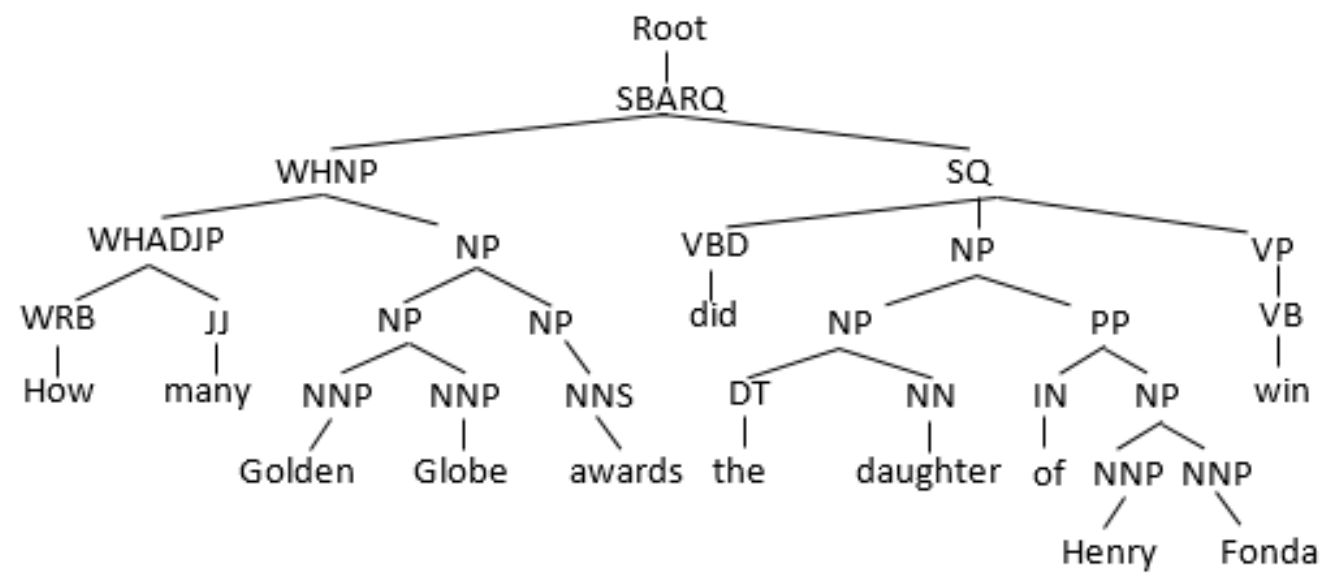}
		\caption{The structural tree of Q3.}
		\label{fig:SP}
	\end{figure}

	We formally define the concept of composite question dealt with in this paper as follows:
	
	\begin{definition}[Composite Question]
		\label{def:Composite Question}
		A composite question \texttt{cq} is a long natural language question (i.e., containing several clauses), which can be decomposed into multiple individual sub-questions denoted by $q_i$.
		In a given \texttt{cq},  sub-questions are connected to each other via an operator $\theta_i$ from the set $\Theta = \{\cap , \cup, \uparrow, F\}$. The operator $\cap$ stands for intersection, $\cup$ for union, $\uparrow$ for assignment and $F$ for function. These operators are derived, respectively, from linguistic patterns such as \texttt{and}, \texttt{or}, \texttt{how many}\footnote{These operators will be discussed in more details.}.
		Thus, a composite question can be represented as a sequence of sub-questions with connecting operators in between i.e., $cp=q_1\theta_1 q_2\theta_2...q_{n-1}\theta_{n-1} q_n$.
	\end{definition}
	A sub-question (from an atomic perspective) typically has a triple structure (i.e., \\ \texttt{subject-predicate-object}) abbreviated as $(s,p,o)$\footnote{This definition inspired by RDF triple specified in W3C RDF recommendation \url{https://www.w3.org/TR/rdf-concepts/\#section-Graph-syntax}.}.
	Subjects commonly are a Named Entity and objects are either a Named Entity or literal associated with the subject \cite{decker2000}. The predicate $p$ is a relation between $s$ and $o$.
	The initial contribution of this paper is \emph{organizing possible relations, which can be extracted from natural language questions via dependency and structural trees of the given questions}.
	It should be noted that the dependency tree is the basis for our relation extraction. \\
	Figure \ref{fig:DP} illustrates the dependency tree of Q3. As it is shown, each labeled edge relates two tokens to each other by a \emph{syntactic relation}. For example, the edge labeled \texttt{nsubj} (subject of) relates the verb \texttt{win} to the noun \texttt{daughter}.
	Since the topology and vocabulary of background knowledge graphs are heterogeneously declared in various domains,
	thus, the direction of the extracted relation depends on the background schema or ontology.
	Therefore, herein, in order to have a generic model, we consider both directions of an extracted relation.  
	\texttt{daughter}$\rightarrow$\texttt{win} or \texttt{win}$\leftarrow$\texttt{daughter} (\texttt{daughter} can be either the subject or the object of the relation \texttt{win}, the correct form depends on the background schema).\\	
	Typically, the tokens with the part-of-speech (POS) tag \texttt{noun} are extended to a noun phrase. But in this paper, we rely on structural parser tree to form noun phrases in a minimal manner. Thus, a noun is extended only in the following three cases (i) the dependent tokens (tokens inside the noun phrase) are placed at the same level in the structural tree, (ii) the whole of the noun phrase is situated in quotation marks or is an expression and (iii) the whole of the noun phrase is recognized as a Named Entity. For example, in Q3,   \texttt{"Henry Fonda"} is a Named Entity. \autoref{fig:SP} illustrates the structural tree of Q3, in which the token \texttt{daughter} is not extended because the remaining tokens are at a lower level of the hierarchy.
	
	\section{Schema for Relation Extraction }
	\label{sec:re}
	In the proposed approach, the relations are categorized into six types as briefly represented in Table \ref{tab:TriplePatterns}. 
	Four of them, (i.e. (i) Direct Verbal, (ii) Genitive \& Preposition, (iii) Possessive Adjective+ Whose, and (iv) Appositive), are adopted from the state-of-the-art with applying appropriate modifications and improvements. 
	Furthermore, we added the other two ones (i.e. (i) Noun Phrase and (ii) Comparative or Superlative), which are inspired from our observation over free textual corpus %\cite{sandhaus2008new},
	QALD benchmark \cite{unger20166th} and the English grammar \cite{huddleston2002cambridge,jespersen2003essentials,zandvoort2001handbook}. 
	
	\begin{table}[hpt] 
		
		\caption {The generated triple patterns based on the recognized linguistic patterns. Used indexes are for:($c$) copular verb and ($nc$) for other verbs except copular verbs.}
		%\begin{center}
		\vspace{2mm}
		\centering
		\scriptsize
		\begin{tabularx}{\textwidth}{|l|l|l|X|}
			\toprule
			\centering
			
			\textbf{Category} & \textbf{Linguistic Patterns} & \textbf{Example} & \textbf{Triple Patterns} \\
			
			\midrule                                                                                 
			
			\multirow{15}{1.35cm}{\textbf{ Verbal}} 
			& \multirow{5}{3.4cm}{P1=(\allowbreak{How$\rightarrow$many$\rightarrow$NP1$\rightarrow$VP$\leftarrow$ NP2})}&\multirow{5}{3.3cm}{E1= (How$\rightarrow$many$\rightarrow$ children$\rightarrow$have$\leftarrow$the actor)} &{T1=(NP2, NP1, ?n)}\\
			%\cline{3-3}
			&&&{T2=(NP2, NP1, ?o)}\\
			&&&\textbf{T3=(?s, NP1, NP2)}\\
			&&&{T4=(NP2, VP, NP1)}\\
			&&&{T5=(NP1, VP, NP2)}\\
			\cline{2-4}
			& \multirow{1}{3.4cm}{P2=(How$\rightarrow$ADJP$\rightarrow$VP$\leftarrow$NP1)} &\multirow{1}{3.3cm}{E2= (How$\rightarrow$tall$\rightarrow$is$\leftarrow$John)} &\textbf{T6=(NP1, ADJP, ?o)}\\

			\cline{2-4}
			& \multirow{3}{3.4cm}{P3=\{NP1$\rightarrow$VP\textsubscript{c}$\leftarrow$NP2, prep,\allowbreak{ NP3\}}} &\multirow{3}{3.3cm}{E3= (person$\rightarrow$was$\leftarrow$vice president, of, Hurry Truman)} &\textbf{T7=(NP1, NP2, NP3)}\\
			%\cline{3-3}
			&&&T8=(NP3, NP2, NP1)\\&&&T9=(NP1, VP\textsubscript{c}, NP2)\\
			
			\cline{2-4}
			& \multirow{6}{3.4cm}{P4=\{NP1$\rightarrow$VP\textsubscript{nc}$\leftarrow$NP2, prep,\allowbreak{ NP3\}}} &\multirow{6}{3.3cm}{E4= (actor$\rightarrow$played$\leftarrow$Dan White, in, Milk)} &\textbf{T10=(NP1, VP\textsubscript{nc}, NP2)}\\
			&&&T11=(NP2, VP\textsubscript{nc}, NP1)\\
			%\cline{3-3}
			&&&\textbf{T12=(NP2, VP\textsubscript{nc}, NP3)}\\
			&&&T13=(NP3, VP\textsubscript{nc}, NP2)\\
			&&&[\textbf{T14=(NP1, VP, NP3)}]\\
			&&& [T15=(NP3, VP, NP1)]\\
			\cline{2-4}
			
			& \multirow{2}{3.4cm}{P5=(NP1$\rightarrow$VP$\leftarrow$NP2)} & \multirow{2}{3.3cm}{E5=(Actor$\rightarrow$played$\leftarrow$Dan White)} &\textbf{T16=(NP1, VP, NP2)}\\&&& T17=(NP2, VP, NP1)\\
			\hline
			\multirow{10}{1.35cm}{\textbf{Possessive Adjective\\+\\ Whose}}  &\multirow{2}{3.4cm}{P6=(PRON\textsubscript{ref2NP3}, NP1, VP\textsubscript{c}, NP2)} &  \multirow{2}{3.3cm}{E6=(whose, death, was, accident)} &\textbf{T18=(NP3, NP1, NP2)}\\ &&& T19=(NP2, NP1, NP3)\\ 
			\cline{2-4}
			&\multirow{8}{3.4cm}{P7=(PRON\textsubscript{ref2NP3}, NP1[, VP\textsubscript{nc}, NP2])} &  \multirow{8}{3.3cm}{E7= (his daughter, win, award)} &\textbf{T20=(NP3, NP1, ?o)}\\ &&& T21=(NP3, ?p, NP1)\\ &&& \textbf{[T22=(?o, VP\textsubscript{nc}, NP2)]}\\ &&&{[T23=(NP2, VP\textsubscript{nc}, ?o)]}\\&&& {T24=(NP1, ?p, NP3)}\\&&& {T25=(?s, NP1, NP3)}\\&&& {[T26=(?s, VP\textsubscript{nc}, NP2)]}\\&&& {[T27=(NP2, VP\textsubscript{nc}, ?s)]}\\
			\cline{2-4}

			\hline
			
			\multirow{8}{1.35cm}{\textbf{Noun Phrase}}  
			&\multirow{8}{3.4cm}{P8\allowbreak{=(            E1, E2)}}

			& \multirow{8}{3.3cm}
			{E8= (G8, country)\\E9= (Apple, co-founder)\\E10= (German, mathematicians)} 
			
			& \textbf{T28=(E1,?p, E2)}\\ &&& T29=(E2, ?p, E1)\\ &&&
			T30=(E1, E2, ?o)\\ &&& 
			T31=(?s, E2, E1)\\ &&& 
			T32=(?s, ?p, E1)\\&&& 
			T33=(E1, ?p, ?o)\\&&& 
			T34=(?s, ?p, E2)\\&&&
			T35=(E2, ?p, ?o)\\ 
			\cline{2-4}
			\hline
			\multirow{4}{1.35cm}{\textbf{Genitive\\ \&\\ Preposition} }  &\multirow{4}{3.4cm}{P9=(NP1, prep, NP2)\\P10=(NP2, gen, NP1)} 
			& \multirow{4}{3.3cm}
			{E11= (daughter, of, Obama)\\
				E12= (city, in, Germany)\\
				E13= (use, of, atomic weapon)} &\textbf{T36=(?s, NP1,NP2)}\\&&&{T37=(NP2, NP1, ?o)}\\ &&& \textbf{T38=(NP1, ?p, NP2)}\\&&& T39=(NP2, ?p, NP1) \\ 
			\hline 
			\multirow{2}{1.35cm}{\textbf{Appositive}} &\multirow{2}{3.4cm}{P11=(NP1, appos, NP2)} 
			& \multirow{2}{3.3cm}{E14= (Nordstrom Inc., the retail chain)} & \textbf{T40=(NP1, ?p, NP2)}\\&&& T41=(NP2, ?p, NP1)\\ 
			\hline
			\multirow{5}{1.35cm}{\textbf{Comparative\\ or\\ Superlative} }
			&\multirow{1}{3.4cm}{P12=(NP, nADJ, N)} & \multirow{1}{3.3cm}{E15=(lake,deeper,100)} &\multirow{1}{*}{\textbf{T42=(NP, nADJ, ?n)}}\\
			\cline{2-4}
			&\multirow{2}{3.4cm}{P13=(NP1, [as] nADJ [as], NP2)} & \multirow{2}{3.3cm}{E16= (lake, deeper, 	Lake Baikal)} &{\textbf{T43=(NP1, nADJ, ?n1)}}\\&&&{\textbf{T44=(NP2, nADJ, ?n2)}}\\
			\cline{2-4}
			&\multirow{1}{3.4cm}{P14=(NP1,[as]qADJ[as],NP2)} & \multirow{1}{3.3cm}{E17=(person,stronger,John)} &{\textbf{T45=(NP1,qADJ,NP2)}}\\
			\cline{2-4}
			&\multirow{1}{3.4cm}{P15=(sADJ, NP1)}& \multirow{1}{3.3cm}{E18=(tallest building)} &\textbf{T46=(NP1, sADJ, ?n)}\\
			
			\bottomrule
		\end{tabularx}
		%\end{center}
		\label{tab:TriplePatterns} 
	\end{table}
	\begin{comment}
	\begin{figure}[ht]
	\centering
	\includegraphics[width =1\textwidth]{Images/grammarElements.PNG}
	\caption{Grammar elements discussed in \cite{EnglishGrammar}}
	\label{fig:EnglishGrammar}
	\end{figure} 
	\end{comment} 
	
	\begin{comment}
	\begin{figure}[ht]
	\centering
	\includegraphics[width =1\textwidth]{Images/myRelations.PNG}
	\caption{Different probable defined relations based on main elements of the grammar of the English language, their subsets and various states of appearing main elements in a clause. }
	\label{fig:myRelations}
	\end{figure}
	\end{comment}
	\begin{enumerate}
		\item \textbf{Verbal Relation:}
		This type of relation is identified when a verbal phrase \texttt{VP} connects two either noun phrases \texttt{NP1}, \texttt{NP2} or a noun phrase \texttt{NP} and an adjective phrase \texttt{ADJP} by a syntactic relation. This relation can be identified via the following patterns. Note that in all of the patterns the adverb modifier is added to the \texttt{VP}, because sometimes it holds a valuable information, which is useful for answering the question, e.g, for the given question \texttt{How old was Steve Jobs sister when she first met him?}, the two adverb modifiers of the verb \texttt{met} are (i) \texttt{when} and (ii) \texttt{first}.
		
		\begin{itemize}
			\item{\emph{Patterng 1:}} When the syntactic pattern \texttt{\texttt{How}$\rightarrow$\texttt{many}$\rightarrow$\texttt{NP1}$\rightarrow$\texttt{VP}$\leftarrow$\texttt{NP2}} is recognized, then the triple patterns T1, T2, T3, T4 and T5 of Table \ref{tab:TriplePatterns} are possibly applicable. In case of T1 and T2, the \texttt{NP2} and \texttt{NP1} are respectively placed as the subject $s$ and predicate $p$. The object $o$ is a variable with the type of either literal value (e.g., numerical ) $?n$ or entity $?o$. 
			The triple pattern T3 is the opposite pattern of T2. In case of T4 and T5 the \texttt{NP1} and \texttt{NP2} are placed as the subject $s$ or object $o$ and the \texttt{VP} is placed as predicate $p$.  
			%    The operator of \texttt{count} is required to be run on the result set of T2 and T3 (cf. Figures \ref{fig:count1.PNG} and \ref{fig:count2.PNG}). 
			Regarding Q3, for the syntactic pattern \texttt{How$\rightarrow$many$\rightarrow$Golden Glob awards$\rightarrow$win$\leftarrow$daughter}, the following triple patterns are generated: (i) \texttt{T1= (daughter, Golden Glob awards, ?n)}, (ii) \texttt{T2= (daughter, Golden Glob awards, ?o)}, (iii) \texttt{T3= (?s, Golden Glob awards, daughter)}, (iv) \texttt{T4= (daughter, win, Golden Glob awards)} and (v) \texttt{T5= (Golden Glob awards, win, daughter)}.    
			\item \emph{Pattern 2:} In the case of identifying the syntactic pattern  \texttt{How}$\rightarrow$ \texttt{ADJP}$\rightarrow$ \texttt{VP}$\leftarrow$ \texttt{NP1},
			the triple pattern T6 from \autoref{tab:TriplePatterns} is generated,
			e.g., for the given question \texttt{How tall is John?}, the triple  \texttt{T6=(John, tall, ?o)} is generated. 
			\item \emph{Pattern 3 and 4:} In the case of the pattern \allowbreak{ (\texttt{NP1}$\rightarrow$\texttt{VP}$\leftarrow$\texttt{NP2}, \texttt{prep}, \texttt{NP3})}, the following two situations (depending whether \texttt{VP} is a copular verb\footnote{A copular verb is a special kind of verb used to join an adjective or noun complement to a subject.}) are considered:\\
			\textbf{\texttt{VP} is a copular verb}:
			In this case, three triple patterns are applicable as (i) \texttt{T7=(NP1, NP2, NP3)} and (ii) \texttt{T8=(NP3, NP2, NP1)}, and (iii) \texttt{T9=(NP1, VP, NP2)}. With respect to Q1, the identified pattern is: \texttt{\{Who$\rightarrow$was$\leftarrow$vice president, under, president\}}, thus the triple patterns (i) \texttt{(Who, vice president, president)}, (ii) \texttt{(president, vice president, Who)} and (iii)  \texttt{(Who, was, vice president)}  are generated.\\ 
			\textbf{\texttt{VP} is not a copular verb}:
			Here, two kinds of indirect verbal relations;  passive and active (as discussed below), are generated, and two triple patterns,  T10 and T11, are usually used. T10 and T11 consume \texttt{NP1} and  \texttt{NP2}. Considering Q2, the two generated triple patterns are (i) \texttt{(actor, plays, Dan White)}, (ii) \texttt{(Dan White, plays, actor)}.\\ \emph{Indirect Passive Verbal Relation:} In this case, four triple patterns i.e., T10, T11, T12, T13 are generated. T12 and T13 consume \texttt{NP2} and  \texttt{NP3} (in case of a passive relation).
			%and place them in positions of subject $s$ and object $o$. Four triple patterns place \texttt{VP} in position of predicate $p$. 
			With respect to Q2, the two generated triple patterns are (i) \texttt{(Dan White, plays, Milk)} and (ii) \texttt{(Milk, plays, Dan White)}.\\  
			\emph{Indirect Active Verbal Relation: }
			When  \texttt{NP2} and \texttt{NP3} are named entities, numbers or embedded inside quotations and \texttt{VP} is non-copular  verb, the triple patterns T14-T15 are generated. With respect to the example question Q2, the pattern  \texttt{\{actor$\rightarrow$plays$\leftarrow$Dan White, in, Milk\}} is recognized as an instance of the Indirect Passive Verbal and Indirect Active Verbal relations. A few of the generated triple patterns are represented in Figure \ref{fig:VPVP}.
			\item{\emph{Pattern 5:}} In the remaining cases, the two noun phrases are placed as subject $s$ and object $o$, and the verbal phrase is placed as the predicate $p$ of the triple, i.e., T16, T17. Considering Q1, the two triples (i) \texttt{(president, approved, Japan)} and (ii) \texttt{(Japan, approved, president)} are generated.
			
		\end{itemize}
		
		\begin{figure}[t]
			\centering
			\includegraphics[scale=0.55]{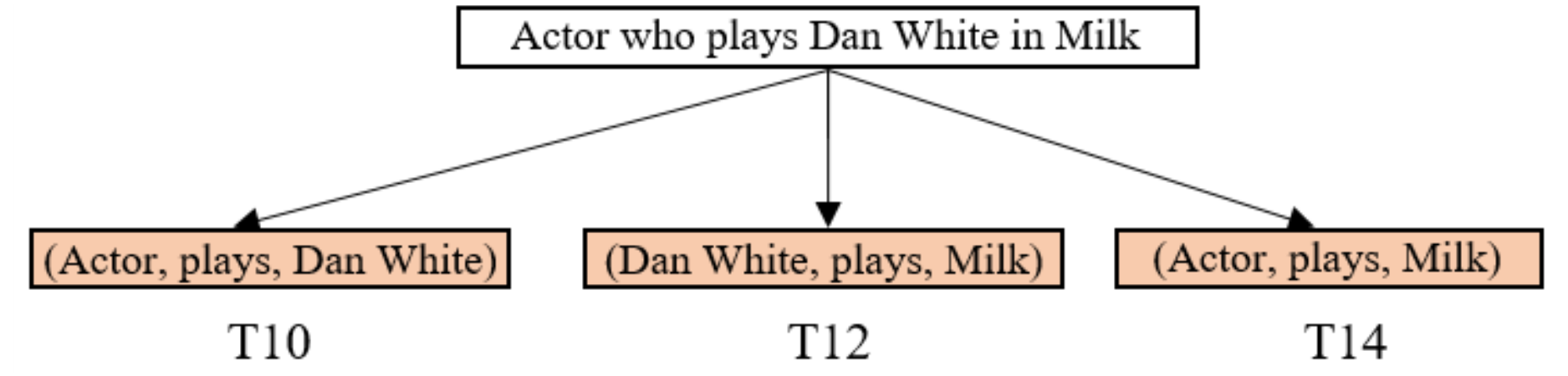}
			\caption{A few of triple patterns generated from the recognition of the P4 linguistic pattern for \texttt{``actor who plays Dan White in Milk''}.}
			
			\label{fig:VPVP}
		\end{figure}
		
		\item \textbf{Possessive Adjective + Whose Relation:} This type of relation is recognized when a possessive adjective or \texttt{whose} (denoted as \texttt{PRON}) appears in the given composite question by the pattern i.e., \texttt{PRON}, \texttt{NP1} \{$\rightarrow$\texttt{VP}$\leftarrow$\texttt{NP2}\}, where \texttt{PRON} refers to a noun phrase \texttt{NP3}, denoted as ref2NP3, and the presence of  \{\texttt{$\rightarrow$VP$\leftarrow$NP2}\} is optional. If the optional pattern \{\texttt{$\rightarrow$VP$\leftarrow$NP2}\} is present, then the triple patterns  T20, T21, T24 and T25 are generated, otherwise, in case \texttt{VP} is a copular verb, T18 and T19, and in the remaining cases, T20-T27 are generated.
		Regarding the question \texttt{Of the people that died of radiation in Los Alamos, whose death was an accident?}, since the pattern \texttt{whose, death, was, accident} is recognized and \texttt{whose} refers to \texttt{people} and the verb \texttt{was} is a copular verb, thus the triple patterns \texttt{(people, death, accident)} and \texttt{(accident, death, people)} are generated. 
		
		\item \textbf{Noun Phrase Relation:} This type of relation is recognized when a sequence of adjective phrases \{\texttt{ADJP1}, \texttt{ADJP2}, ..., \texttt{ADJPn}\} is preceded a sequence of noun phrases \{\texttt{NP1}, \texttt{NP2}, ...,\texttt{NPn}\}, denoted by \texttt{NP} in a given composite question (the presence of adjective phrases is optional).
		This pattern is transformed into a set of \textbf{paired words} using the following two proposed approaches:
		\begin{itemize}
			\item{\emph{Noun phrase extraction I:}} The nearby words are paired. 
			For instance, considering the noun phrase pattern \texttt{`human brain function'}, the two paired words are (i) \texttt{`human brain'} and (ii)  \texttt{`brain function'}.
			
			\item{\emph{Noun phrase extraction II:}} Pairing  based on the neighboring policy is limited only to the noun phrases, whereas an adjective is paired with the right most noun phrase in \texttt{NP}. For example, in the given pattern \emph{`10-year Japanese government bond'}, \texttt{`10-year'} and \texttt{`Japanese'} are adjectives and \texttt{`government'} and \texttt{`bond'} are noun phrases. Thus, the three pairs are: (i) \texttt{`10-year bond'}, (ii) \texttt{`Japanese bond'} and (iii) \texttt{`government bond'}.
			
		\end{itemize}
		
		After extracting paired words, we have to map them to the underlying background knowledge graph. Note that for a given paired words, we refer to the first word, which is either an adjective or a noun by \texttt{E1}, and the second word, which is a noun, by \texttt{E2}.
		Table \ref{tab:TriplePatterns} lists all the possible triples patterns, i.e., T28 to T35, which can be generated through the following conditions:
		
		\begin{itemize}
			
			\item{\emph{Pattern 1:}} 
			When \texttt{E1} can be mapped into an entity in the underlying knowledge graph, but not  \texttt{E2}, then there are two states: (I) \texttt{E2} is mappable to a class from the underlying ontology (or schema of the underlying knowledge graph), (II) \texttt{E2} is not mappable to a class.
			In the first state, the triple patterns T28, and T29 are generated merely by placing \texttt{E1} and \texttt{E2}, respectively, in the subject or object positions. A variable $?p$ is situated in the predicate position since it is unknown, e.g., in the given question \texttt{``Which Chinese-language country is a former Portuguese colony?''}, the words \texttt{Chinese-language}, \texttt{E1}, and \texttt{country}, \texttt{E2}, are paired.
			The former one is an NP, and furthermore, the word is mappable to the entity \texttt{dbr:Chinese\_language}\footnote{http://dbpedia.org/resource/Chinese\_language} in DBpedia \footnote{http://wiki.dbpedia.org/} knowledge graph while the latter one is not (although is mappable to a class in ontology).
			Accordingly, the triple patterns (i) {(\texttt{country,?p,Chinese-language})} and (ii) {(\texttt{Chinese-language, ?p, country}) are generated}.\\		In the second state, (which \texttt{E2} is not mappable to a class), the triple pattern T30 and T31 are generated, where \texttt{E2} is placed as the predicate $p$ and \texttt{E1} is placed as subject and object, e.g., in question \texttt{``Who is the child of Apple co-founder?''}, the two words \texttt{`Apple'} and \texttt{`co-founder'} are paired, whereas the former one is mappable to the entity \texttt{dbp:Apple\_Inc.}\footnote{http://dbpedia.org/page/Apple\_Inc.} of DBpedia and the latter one is not mappable to neither an entity nor a class, thus the generated triple patterns are \texttt{(?s,co-founder,Apple)} and \\ \texttt{(Apple,co-founder,?o)}.\\	
			\item{\emph{Pattern 2:}} When the both \texttt{E1} and \texttt{E2} are mappable to entities of the underlying knowledge graph; then, the four triple patterns T32-T35 of Table \ref{tab:TriplePatterns} are generated, e.g., in the question \texttt{``Which German mathematicians were members of the von Braun rocket group?''}, \texttt{German} and \texttt{mathematicians} are mapped as two different named entities \texttt{dbr:Germany}\footnote{<http://dbpedia.org/resource/Germany>} and \texttt{dbr:Mathematics}\footnote{<http://dbpedia.org/resource/Mathematics>} respectively.

			\item{\emph{Pattern 3:}} In the remaining cases, the triple patterns T28-T35 from Table \ref{tab:TriplePatterns} are generated (e.g., \texttt{`steady growth'}, \texttt{`first wife'}). 
			
		\end{itemize}
		
		\item
		\textbf{Genitive \& Preposition Relation:} This relation is recognized when a preposition \emph{prep} is identified between the two noun phrases \texttt{NP1} and \texttt{NP2} \allowbreak{(i.e. \texttt{NP1}, \texttt{prep}, \texttt{NP2}}) (they are not necessary adjacent \allowbreak{e.g., \texttt{NP3}$\rightarrow$\texttt{VP}$\leftarrow$\texttt{NP1}, \texttt{prep}, \texttt{NP2}}, when \texttt{VP} is copular verb).
		Generally, the four triple patterns T36-T39 from Table \ref{tab:TriplePatterns} are generated.
		There is an exception as follows(we named this exception as Genitive \& Preposition reformation):
		In case of the patterns \allowbreak{(\texttt{NP1}$\rightarrow$\texttt{VP}$\leftarrow$\texttt{NP2}, \texttt{prep}, \texttt{NP3}}) and \allowbreak{(\texttt{NP1}, \texttt{prep\textsubscript{1}}, \texttt{NP2}, \texttt{prep\textsubscript{2}}, \texttt{NP3})}, if \texttt{NP2} and  \texttt{NP3} are named entities (mapped to entities), numbers or surrounded by quotations, 
		then the Genitive \& Preposition relation between \texttt{NP2} and \texttt{NP3} is ignored and 
		in case of the former pattern only an Indirect Verbal relation between  \texttt{NP1} and \texttt{NP3} is considered while in case of the latter pattern, a Genitive \& Preposition relation between  \texttt{NP1} and \texttt{NP3} is considered, e.g., for the given sentence \texttt{``During the Punitive Expedition into Mexico in 1916''}, the two Genitive \& Preposition relations are (i) \texttt{into} between \texttt{`Punitive Expedition'} and \texttt{`Mexico'} and (ii) \texttt{in} between \texttt{`Mexico'} and \texttt{`1916'}. Since \texttt{`Mexico'} is a Named Entity and \texttt{`1916'} is a number, thus the relation \texttt{in} is ignored and a Genitive \& Preposition relation between \emph{`Punitive Expedition'} and \emph{`1916'} is considered. 
		Furthermore, one of nexus-substantives in English language grammar reported in \cite{jespersen2003essentials} is genitive (possessive \emph{s} pronoun or preposition \emph{of}) denoted by \emph{gen}, then we encounter a sequence \texttt{(NP2, gen, NP1)} like a prepositional relation \texttt{(NP1, prep, NP2)}. For considering sequence \texttt{(NP1, prep, NP2)} or \texttt{(NP2, gen, NP1)},

		\item \textbf{ Appositive  Relation:}
		This type of relation is recognized when an appositive (\texttt{appos}) appears between the two noun phrases \texttt{NP1} and \texttt{NP2}. Thus, the triple patterns T40 and T41 from Table \ref{tab:TriplePatterns} are generated by placing \texttt{NP1} and \texttt{NP2} as either subject or object. 
		In case of having the pattern \allowbreak{(i.e. \texttt{NP1} $\rightarrow$ \texttt{VP} $\leftarrow$ \texttt{NP2}, \texttt{appos}, \texttt{NP3}}), if a Direct Verbal relation is identified between \texttt{NP1} and \texttt{NP3} \allowbreak{(i.e. \texttt{NP1}$\rightarrow$\texttt{VP}$\leftarrow$\texttt{NP3})}, then the Appositive relation between \texttt{NP2} and \texttt{NP3} is ignored (It is referenced as \texttt{Appositive relation filtering}). Because when a noun phrase \texttt{NP2} has an appositive relation with other noun phrase \texttt{NP3}, semantically the second noun phrase is supported the first one and is not considered as the object for the given verb, e.g., for the given sentence \texttt{``A physician 's diet , exercise , The brewer said the improvement in trading performance was due to increased volumes led by canned Draught Guinness , an 11 percent increase in productivity per employee and returns on a 10 percent increase in marketing investment to 107 million pounds .''}, the pattern (\texttt{increased volumes}$\rightarrow$\texttt{led}$\leftarrow$\texttt{canned Draught Guinness}, appos, \texttt{an 11 percent increase}) and (\texttt{increased volumes}$\rightarrow$\texttt{led}$\leftarrow$\texttt{an 11 percent increase}) is identified, thus the Appositive relation between \texttt{``canned Draught Guinness''} and \texttt{``an 11 percent increase''} is removed.

		\item \textbf{Comparative or Superlative Relation:} This type of relation is recognized when a comparative or superlative clause is identified in the following patterns:
		
		\begin{itemize}

			\item \emph{Pattern 1:} Totally comparative adjectives are from two types for stating quality (e.g., worse, better,...) or quantity (e.g., higher, deeper,...) of a property. Different triple patterns are generated for each of them. To distinguish between these two types in Table \ref{tab:TriplePatterns} the quantity type is named numerical \texttt{nADJ} and the other one is named quality \texttt{qADJ}.  If a 
			\emph{numerical comparative adjective} \texttt{nADJ} is identified between a noun phrase \texttt{NP} and a number \texttt{N} then the triple pattern T42 is generated by placing \texttt{NP}, \texttt{nADJ} and the variable \texttt{?n} respectively in subject, predicate and object positions.
			If \texttt{nADJ} appears between the two noun phrases \texttt{NP1} and \texttt{NP2}, then the triple patterns T43 and T44 are generated.
			%A function \texttt{filter(N,O,?n)} to filter results considering $N$ by the operator $O$ based on \texttt{nADJ} is applied on T42. (see Figure \ref{fig:filter1}). Furthermore, a filtering function  is applied on the both triples to compare $?n1$ with $?n2$.
			When a \emph{quality comparative adjective} \texttt{qADJ} is identified between the noun phrases \texttt{NP1} and \texttt{NP2}, then triple pattern T45 is generated. Please be noted that the bracket \texttt{[as]} is used to consider equality in comparison, e.g., for the given question \texttt{``Are there man-made lakes in Australia that are deeper than 100 meters?''}, the triple pattern \texttt{(lakes, deeper, ?n)} is generated.
			
			\item \emph{Pattern 2:}  If a 
			\emph{quantity superlative adjective} (\texttt{sADJ}) is identified before a noun phrase \texttt{NP}, then the associated triple pattern (i.e., T46) is generated by placing \texttt{NP}, \texttt{sADJ} and the variable \texttt{?n} respectively in subject, predicate and object positions, e.g., for the given question: \texttt{``Who are the architects of the tallest building in Japan?''}, the triple pattern \texttt{(building, tallest, ?n)} is generated.
		\end{itemize}
		%?uri <http://dbpedia.org/ontology/depth> ?num . FILTER(?num > 100)
	\end{enumerate}
	
	\begin{comment}

	\begin{figure}[ht]
	\begin{center}
	\subfloat[]{\includegraphics[width =0.3\textwidth]{Images/count1.PNG}\label{fig:count1.PNG}} 
	\subfloat[]{\includegraphics[width = 0.3\textwidth]{Images/count2.PNG}\label{fig:count2.PNG}}  
	
	\caption{Needed Function \texttt{count} to apply On triple patterns (a) T2 and (b) T3.}
	\end{center} 
	\end{figure}

	\begin{figure}[ht]
	\begin{center}
	\subfloat[]{\includegraphics[width =0.35\textwidth]{Images/filter1.PNG}\label{fig:filter1}} 
	\subfloat[]{\includegraphics[width = 0.63\textwidth]{Images/filter2.PNG}\label{fig:filter2.PNG}}  
	
	\caption{Needed function \texttt{filter} to apply on (a) triple pattern  T40 in and (b) triple patterns T41 and T42.}
	\end{center} 
	\end{figure}

	\begin{figure}[ht]
	\centering
	\includegraphics[width =0.8\textwidth]{Images/Filter.PNG}
	\caption{Needed function \texttt{filter} to apply on triple pattern  T30 in left sub tree and triple patterns T40 and T41 in right sub tree.} \label{fig:Filter}
	\end{figure} 
	\end{comment}

	\section{Aggregating answer set of sub-questions}
	\label{sec:approach}

	To answer composite questions, we initially decompose the given question into a series of sub-questions, which each one is mappable to a triple pattern. 
	Each sub-questions might be either dependent or independent. 
	An independent sub-question (atomic sub-question), can be directly answered, while a dependent sub-question requires collecting the answers from the neighboring sub-questions for the retrieval task. 
	Thus, the order of issuing sub-questions for the retrieval task is crucial.
	It is a more challenging task, where the underlying target data sources are heterogeneous.
	We formally describe this challenge in the following.
	
	\begin{challenge}[Aggregating answer set of sub-questions]
		\label{challenge:Answering the Composite question}
		The final answer of \texttt{cq} is an aggregation of the answer set of sub-questions. In hybrid search space, each sub-question should be searched from one source (structured or unstructured source). Thus, the order of executing sub-questions, as well as aggregating results, is the main challenge. For example, with respect to the given composite question \texttt{`In which city where Charlie Chaplin's half brothers born?'}, we distinguish the two sub-questions $q_1= (?s, Born,city)$ and $q_2= (?s, half brother, Charli Chaplin)$. 
		The former sub-question should be executed after integrating the answer set of the latter one.
		So, (i) the order of executing sub-questions and (ii) the manner of aggregating the answer set of sub-questions are substantial steps for answering the input query in a hybrid search space.
		
	\end{challenge}
	\begin{figure}[ht]
		\centering
		\includegraphics[width =0.9\textwidth]{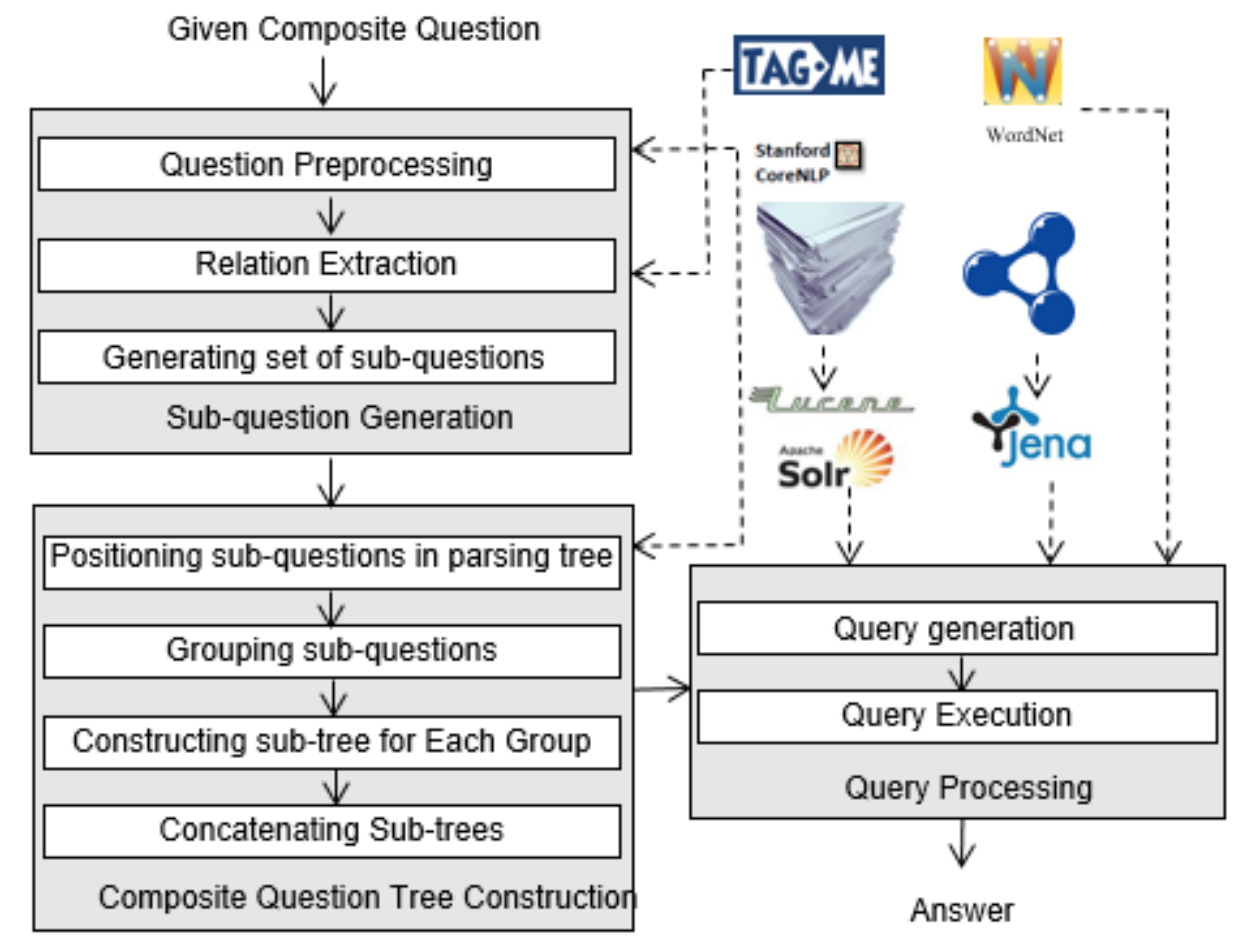}
		\caption{The structure of proposed approach}
		\label{fig:structure.pdf}
	\end{figure}
	In this section, we propose an approach addressing this challenge.
	This approach relies on a user study experiment to find the cognitive procedure of the human mind for answering composite questions as follows: 
	\begin{finding}[Cognitive procedure of human brain]For this experiment, we provided five composite questions for 15 people from various age groups and educational backgrounds.
		They were asked to distinguish the proper ordering for answering a given composite question. 
		This experiment revealed that cognitive procedure of human for ordering sub-questions is in-line with an infix arithmetic expression (also called parenthesized) \cite{miller2013}, where inner sub-expressions are prioritized to outer ones for processing; which confirm our intuition.
	\end{finding}
	Inspired by this observation, our proposed approach relies on prioritizing inner sub-questions when retrieving the answer.
	It is important to mention that the required pre-processings for a given composite question is: (i) dependency parsing, (ii) constituency parsing \cite{Schuster2016}, (iii) part of speech taging \cite{Toutanova2013}, (iv)  Recognition (NER) \cite{isem2013daiber,PaoloFerragina2010}, (v) Named Entity Linking \cite{isem2013daiber,PaoloFerragina2010}, and (vi) stop words removal.

	\paragraph{\textbf{Generating set of sub-questions.}}
	\label{sec:Sub-questions generating}
	We produce the possible sub-question set denoted by $Qs$ for the given composite question employing strategies and triple patterns represented in the previous section (i.e., summarized in Table \ref{tab:TriplePatterns}).
	In this table, there are triple patterns printed in bold (so-called \textbf{key triple patterns}) which follow the same order of tokens in the given composite question.
	Since in our proposed approach, the order of tokens in the given composite question matters; in this step   \emph{key triple patterns} are exclusively taken into account. Other triple patterns will be used in query execution step for query extension. For those relations which more than one \emph{key triple patterns} is defined while containing common place holders (independent of their positions), only one of them is used for sub-question set generation task.
	Each linguistic pattern discussed in Section \ref{sec:problemStatement} is mapped into only one single sub-question using its key triple pattern.
	For the given running example 
	\allowbreak{\texttt{"Which writers had influenced the philosopher that refused a Nobel Prize?"}}, the sub-question set is generated as \allowbreak{
		\texttt{$Qs_1$: 
			\{(writers, influenced, philosopher), (philosopher, refused,  Nobel Prize)\}};}
	
\begin{figure}[ht]
		\centering
		\includegraphics[scale=0.9]{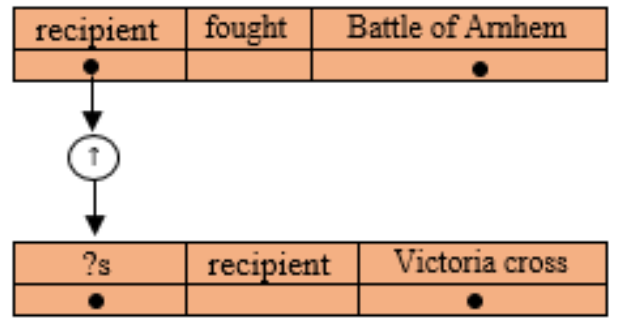}
		\caption{Tree for our running example question Q4}
		\label{fig:recipientTree.PNG}
	\end{figure}

\begin{figure}[ht]
		\centering
		\includegraphics[width =1\textwidth]{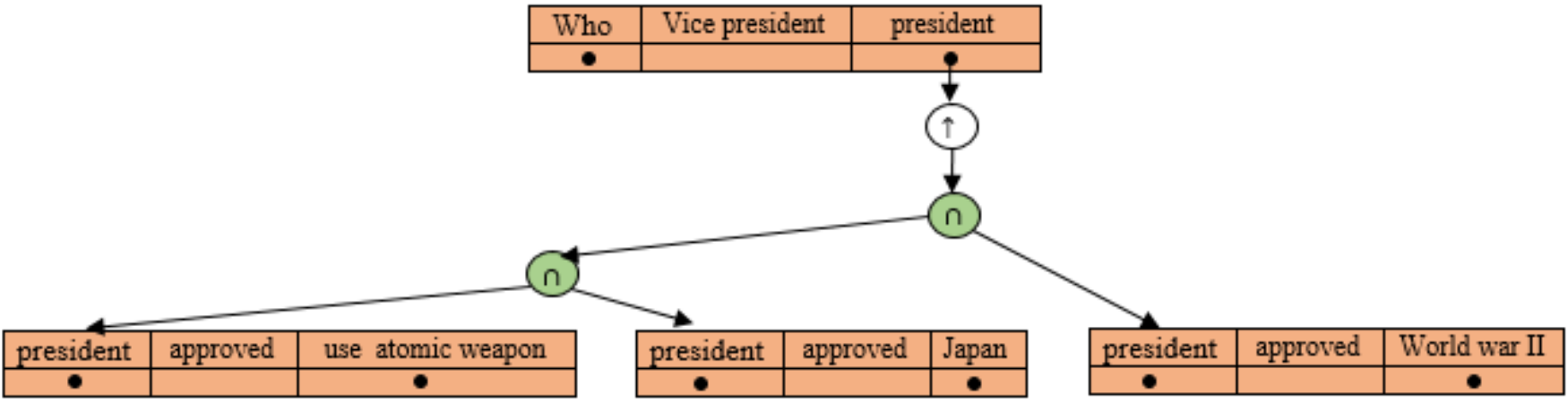}
		\caption{Tree for composite question ``Who is vice president under president who approved atomic weapon against Japan during World War II?''}
		\label{fig:presidentTree.PNG}
	\end{figure}

\begin{figure}[htbp]
		\begin{center}
			\subfloat[]{\includegraphics[width =0.5\textwidth]{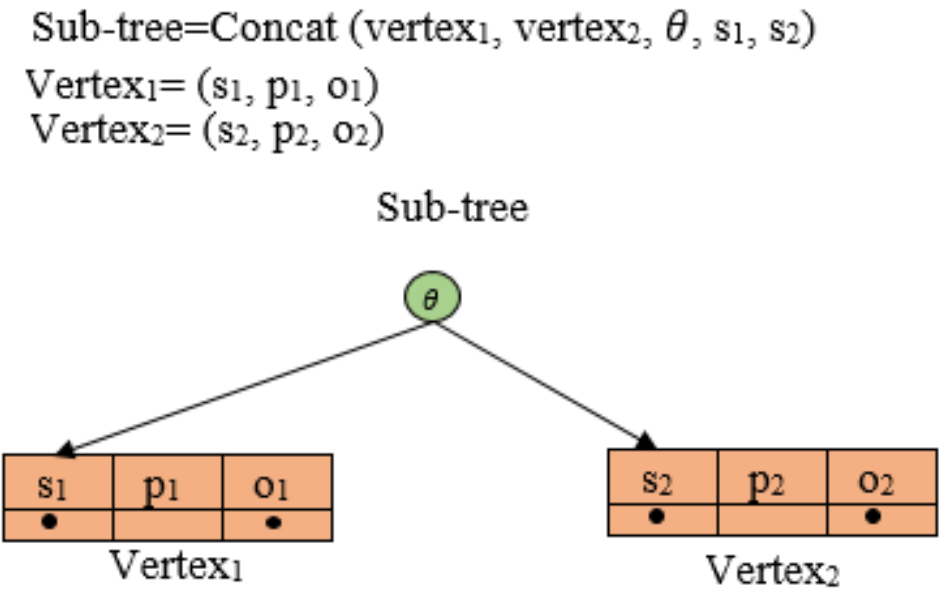}\label{fig:Concat.PNG}} 
			\subfloat[]{\includegraphics[width =0.5\textwidth]{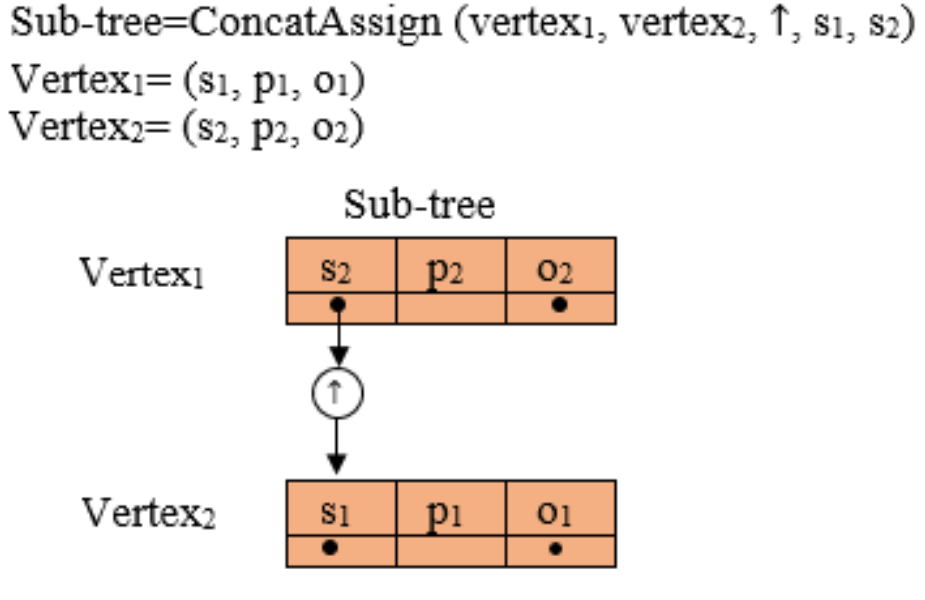}\label{fig:ConcatAssign.PNG}} \\
			\caption{Examples for two functions Concat and ConcatAssign}
			\label{fig:Concates}
		\end{center} 
	\end{figure}

\begin{figure}[hp]
		\begin{center}
			\subfloat[sub-question set]{\includegraphics[width =0.7\textwidth]{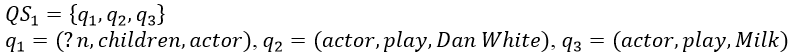}\label{fig:sub-questions.PNG}} \\
			\subfloat[Determining the position of sub-questions in Structural parsing result]{\includegraphics[width =0.9\textwidth]{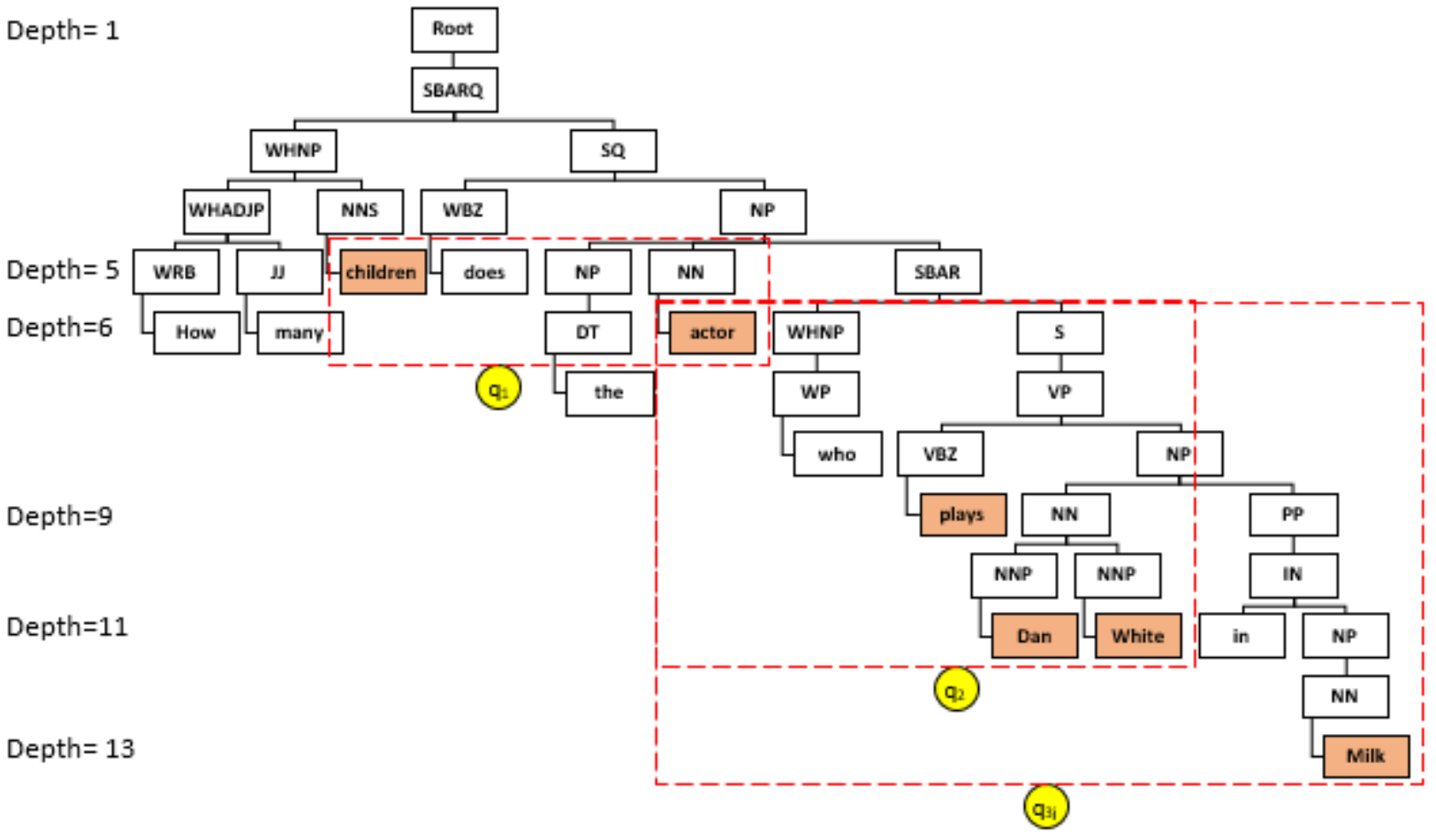}\label{fig:fusion.PNG}} \\
			\subfloat[Structural parsing result]
			{\includegraphics[width=0.9\textwidth]{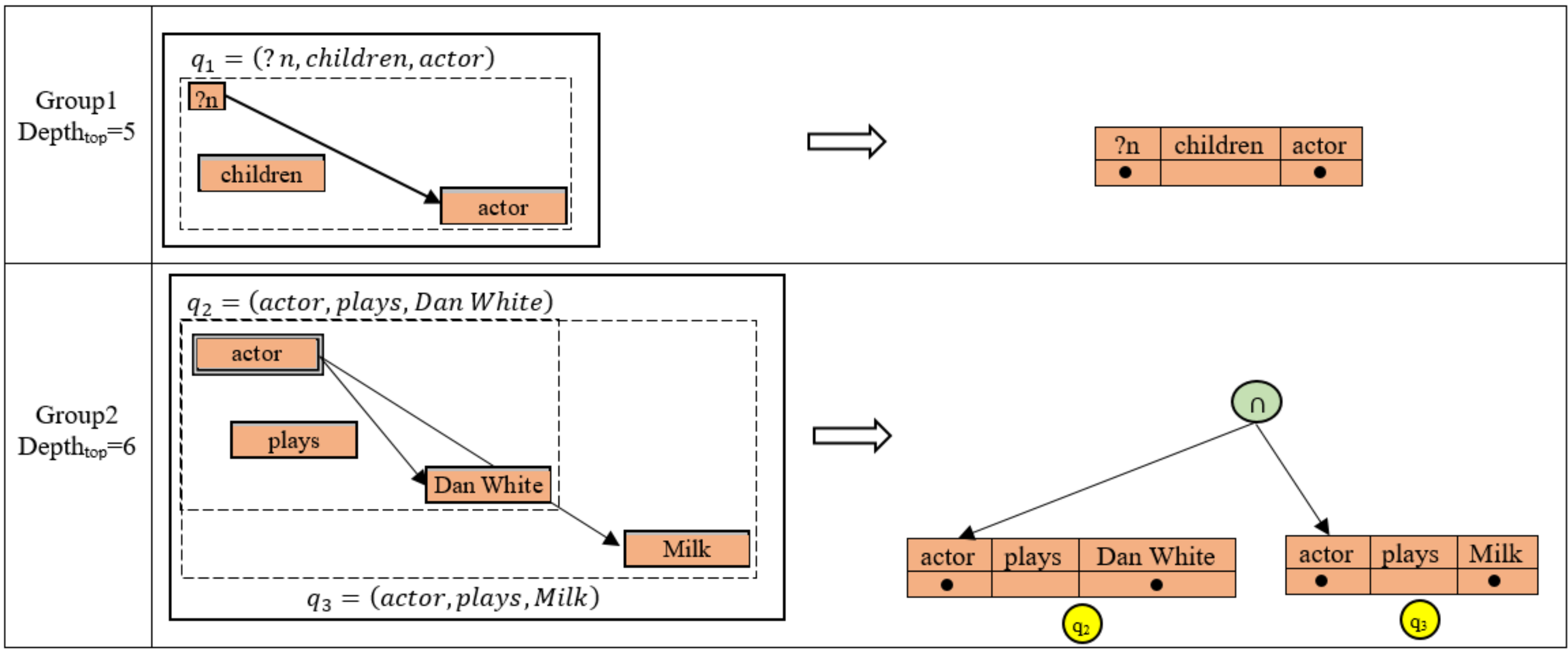}\label{fig:levelsSubTrees.PNG}} \\
			\subfloat[Composite question tree]{\includegraphics[width = 0.6\textwidth]{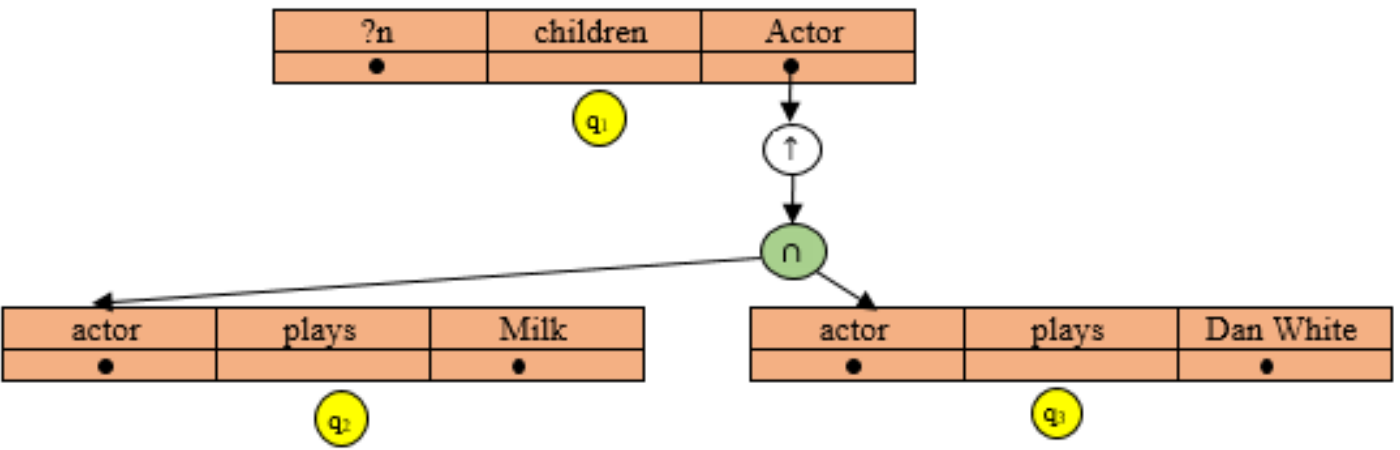}\label{fig:newtree}}  
			
			\caption{Constructing the composite question tree for the example question Q2}
			\label{fig:compositeExample}
		\end{center} 
	\end{figure}

\begin{figure}[!h]
		\begin{center}
			\subfloat[]{\includegraphics[scale=0.7]{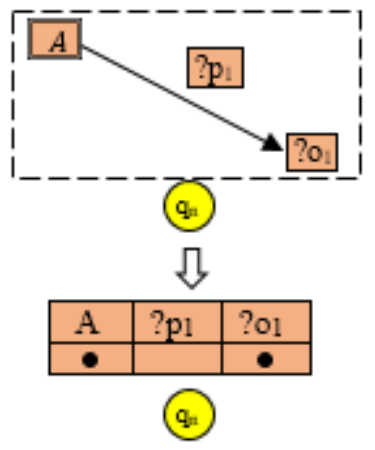}\label{fig:stateSingle.PNG}} 
			\subfloat[]{\includegraphics[scale=0.7]{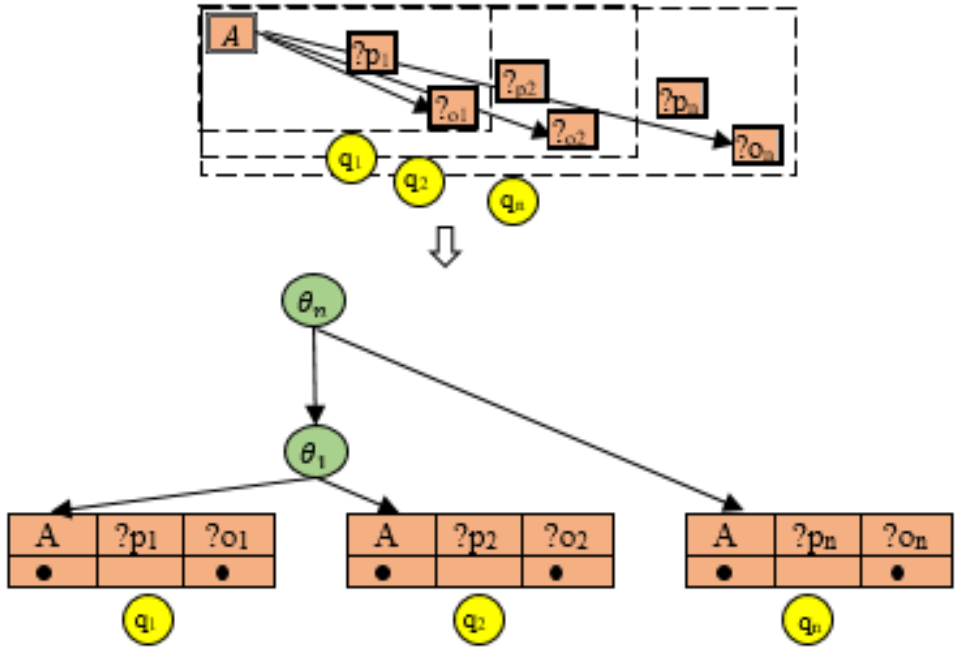}\label{fig:stateS.PNG}} \\
			\subfloat[]{\includegraphics[scale=0.7]{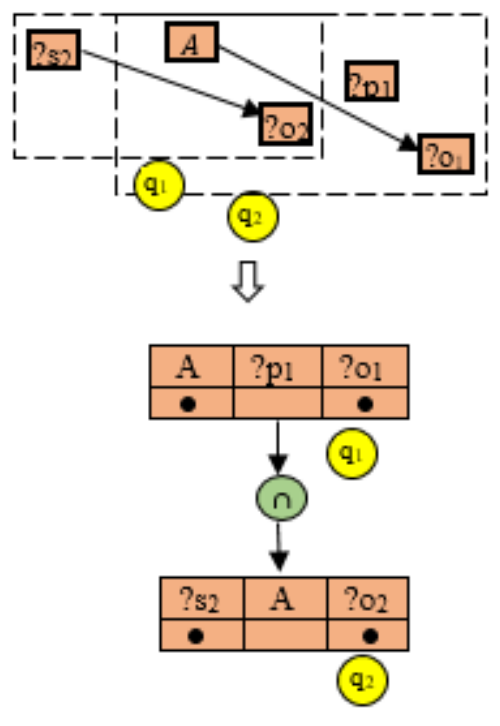}\label{fig:stateP.PNG}} 
			\subfloat[]{\includegraphics[scale=0.7]{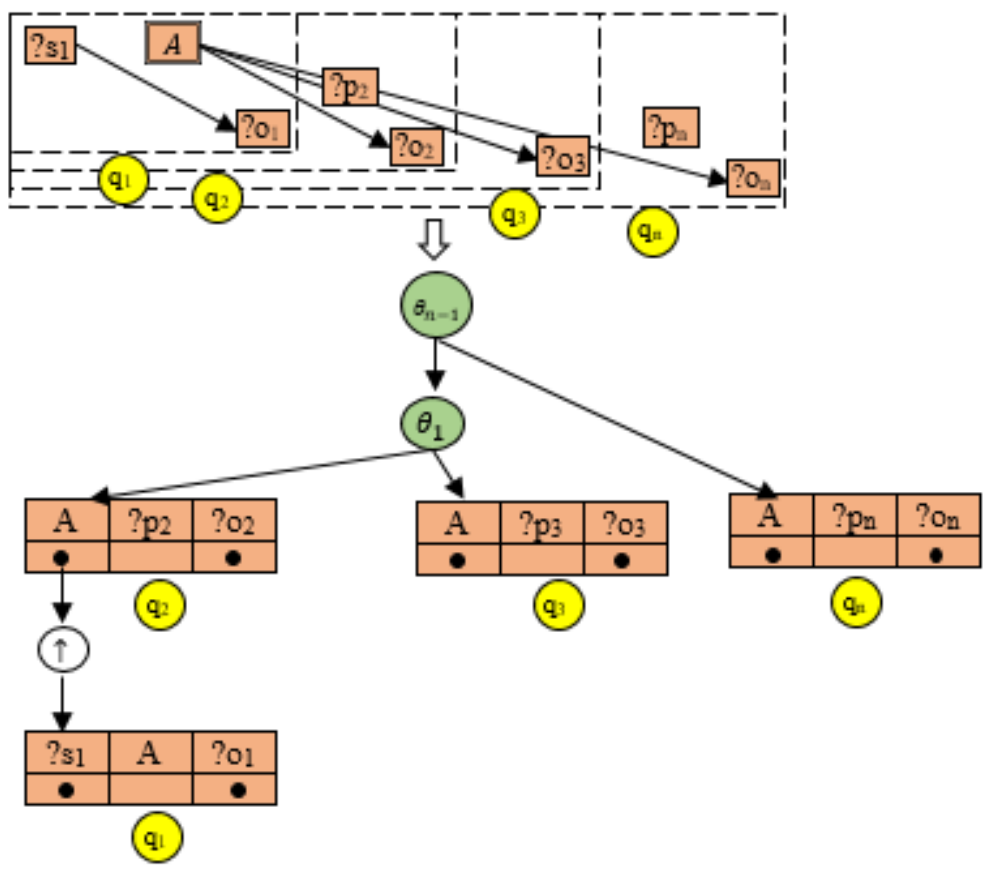}\label{fig:statePS.PNG}} 
			\\

			\caption{Different possible states of sub-questions in structural parsing result and regarding vertices in composite question tree}
			\label{fig:differentState}
		\end{center} 
	\end{figure}

	\paragraph{\textbf{Constructing tree for composite question.}}
	Regarding the challenge \ref{challenge:Answering the Composite question},
	a critical step is 
	\emph{ordering} execution schema and \emph{aggregating} answer sets from all the derived sub-questions. 
	To address this challenge, we propose a novel approach to determine appropriate order of sub-questions of a given composite question for execution and subsequently aggregating answers. 
	This approach relies on a tree structure constructed on the driven sub-questions concerning syntactic and structural parsing (i.e., dependency tags and constituency pars) of the given composite question. A tree of a given composite question is formally defined as (Figure \ref{fig:recipientTree.PNG} and \ref{fig:presidentTree.PNG} shows the constructed trees for two given questions): 
	\begin{definition}[Composite Question Tree]
		\label{def:Composite Question Tree}
		A given composite question $cq$ is represented  by a binary tree $cqt=(V,E)$ containing unlabeled directed edges and the set of vertices $V$ as the union of all sub-questions $Qs$ and operators $\Theta$,  $V= Qs \cup \Theta$ ($\Theta = \{\cap , \cup, \uparrow, F\}$).  The leaves of $cqt$ are always sub-questions while the root can be either a sub-question $q_i$ or an operator from $\Theta$. 
		
	\end{definition}
	Algorithm \ref{alg:ConstituenctTree} elaborates on our approach for constructing trees of composite question. Its inputs are 
	structural parsing tree \cite{Schuster2016} and the set of sub-questions obtained from Section \ref{sec:Sub-questions generating}. 
	Please note that the output of the Stanford parser is transformed in to a structural parsing tree and then is injected to this algorithm as the input.
	The lines 1-3 of Alg. \ref{alg:ConstituenctTree}
	spot the given sub-questions in the structural parsing tree (Fig.\ref{fig:fusion.PNG} shows the position of sub-questions in structural parsing tree for our running example). 
	Next, the matches are grouped using the concept of top depth (Fig.\ref{fig:levelsSubTrees.PNG}, i.e., line 4).
	After that, Algorithm \ref{alg:mapGroupToSubTree} is called which receives a given group (a couple of sub-questions) and constructs a sub-tree out of it. First, in case of more than one sub-question in a group, all sub-questions are sorted using their forward depth in line 1 in Alg. \ref{alg:mapGroupToSubTree}. There are four distinct possibilities for forming sub-trees (illustrated in Fig. \ref{fig:differentState}). The process of constructing sub-tree for each group has a few steps as followings:  %Sub-questions in each group have four probable  which each state has a different mapping to sub-tree. 
	(i) each sub-question is mapped to an individual vertex in the developing composite question tree $cqt$ (the lines 1-3 in Alg. \ref{alg:mapGroupToSubTree}).
	(ii) it starts concatenating vertices based on the shared subject or predicate.
	The method \emph{ConcatAssign} captures two vertices and connects them by an intermediate node showing an operator and the method \emph{Concat} obtains two vertices and connects them by parental node showing an operator.
	Figure \ref{fig:ConcatAssign.PNG} and \ref{fig:Concat.PNG} show the process of concatenation. 
	This process continues till connecting all the vertices
	(the lines 5-16 of Algorithm \ref{alg:mapGroupToSubTree} performs concatenation process) (Fig.\ref{fig:levelsSubTrees.PNG} shows sub-trees for our running example).\\
	Eventually, the \emph{ConcatAssign} method connects all the sub-trees and delivers the connected composite question tree $cqt$ (the lines 8-11 in Algorithm \ref{alg:ConstituenctTree})(The completed composite question tree for our running example is illustrated in Fig.\ref{fig:newtree}).

	\begin{algorithm}
		\SetKwInOut{Input}{Input}
		\SetKwInOut{Output}{Output}
		
		\Input{Structural parsing tree $spt$, sub-questions set $Qs=\{q_1,...,q_{n}\}$.}
		\Output{Composite question $cqt$.}
		
		\For{$i\leftarrow 1$  \KwTo $n$ } 
		{ $top-depth(q_i),down-depth(q_i)\leftarrow$ Trace-spt($q_i$)\; }
		$Group_1,...,Group_m\leftarrow Grouping(top-depth(q_i),...,top-depth(q_n))$\;
		%$sort_the_group\leftarrow Sort(down-depth(group_1),...,top-depth(group_m))$\;
		\For{$groupID\leftarrow 1$  \KwTo $m$ }{$sub-tree_{groupID}\leftarrow
			Algorithm \ref{alg:mapGroupToSubTree}(group_{groupID})$
		}
		$cqt\leftarrow sub-tree_m$\;
		\For{$groupID\leftarrow  m-1$   \KwTo  $1$ }
		{
			$cqt\leftarrow  ConcatAssign(cqt,sub-tree_{groupID},\uparrow,s_{cqt},o_{{sub-tree}_{groupID}})$\; }
		\caption{Constructing tree for composite questions}
		\label{alg:ConstituenctTree}
	\end{algorithm}
	
	\begin{algorithm}
		\SetKwInOut{Input}{Input}
		\SetKwInOut{Output}{Output}

		\Input{A set of sub-questions $Qs=\{q_1,...,q_{k}\}.$ }
		\Output{A sub-tree for composite question.}
		$sorted-sub-questions\leftarrow Sort(down-depth(q_1),...,down-depth(q_k))$\;
		\For{$i\leftarrow1$ $\KwTo$ $k$}{   $vertex\leftarrow Map(q_i)$\;}
		{ \If{ $k \geq 1$}{ 
				\uIf{$Common(predicate,subject)$}
				{
					$sub-tree\leftarrow ConcatAssign(vertex_2,vertex_1, \uparrow, s_2,s_1)$\;
					\While{$vertex$}
					{$sub-tree\leftarrow Concat(sub-tree,vertex, \Theta,s_{sub-tree}, s_{vertex})$\;}
				}
				\ElseIf{$Common(subject)$}{
					$sub-tree\leftarrow Concat(vertex_1,vertex_2, \Theta, s_1,s_2)$\;
					\While{$vertex$}
					{$sub-tree\leftarrow Concat(sub-tree,vertex, \Theta,s_{sub-tree}, s_{vertex})$\;}
				}

			}
			
		}

		\caption{Constructing sub-tree from sub-questions having same upper depth}
		\label{alg:mapGroupToSubTree}
	\end{algorithm}

	\section{Query Generation and Execution} 
	
\begin{figure}[ht]
		\centering
		\includegraphics[width =1\textwidth]{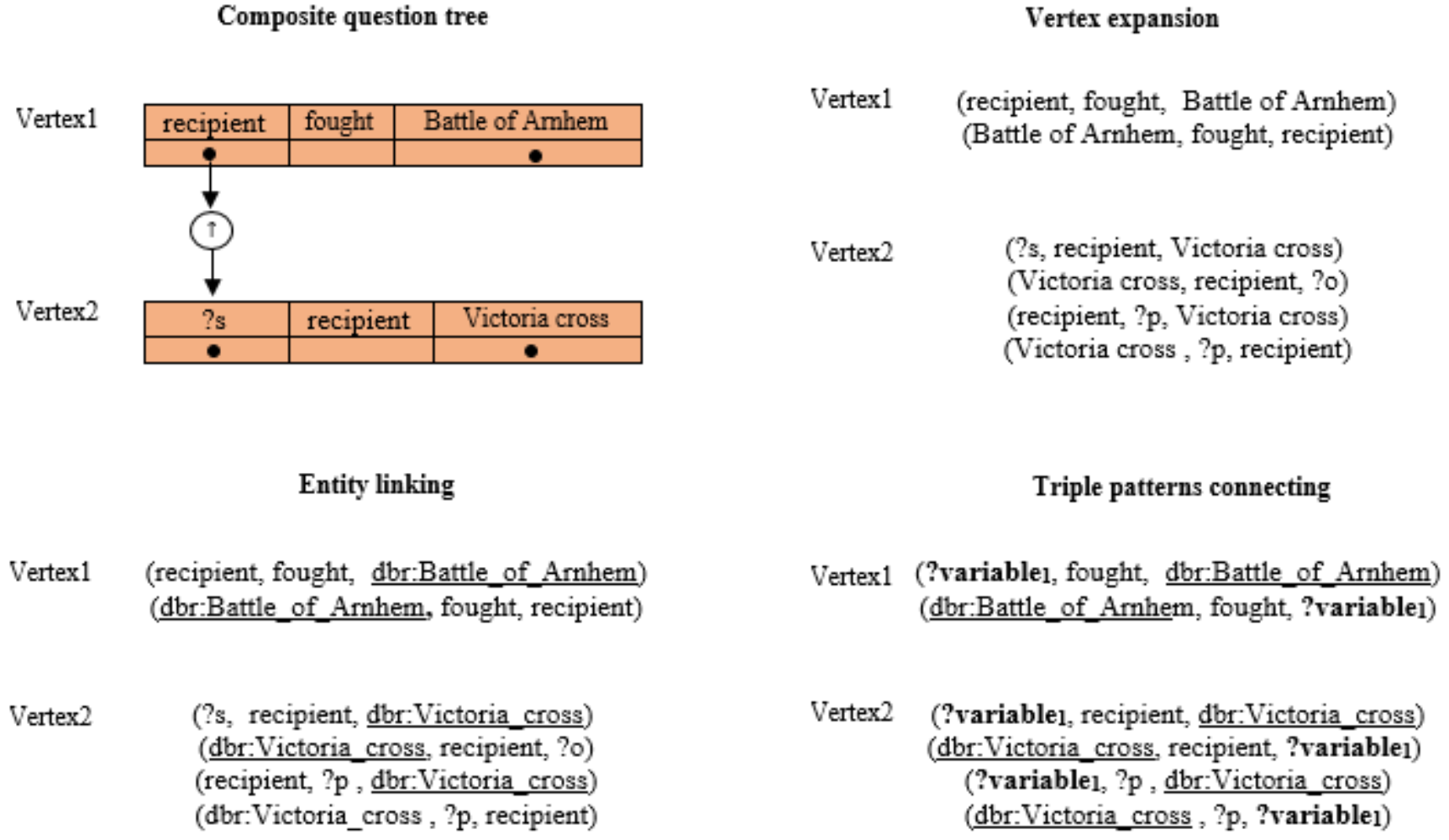}
		\caption{Query Generation}
		\label{fig:query_eneration}
	\end{figure}

	To generate a formal 
	query from a given composite question tree, we have to accomplish the three following tasks (illustrated in Fig.\ref{fig:query_eneration}):
	(i) \emph{vertex expansion:} each vertex of tree is expanded by including all the other relevant triple patterns listed in table \ref{tab:TriplePatterns}.
	(ii) \emph{entity linking:} for a given triple pattern \texttt{(s,p,o)}, each term (either \texttt{s}, \texttt{p} or \texttt{o}) is matched against the underlying knowledge graph, then the candidate matches are placed in the proper positions, in case of linking the object \texttt{o} to a type entity such as \texttt{dbo:City}, a new variable is introduced and placed in the object position, additionally a triple pattern restricting type of the object variable is added (e.g. \texttt{(?variable,rdf:type,dbo:City)}.
	(iii) \emph{predicate expansion:} in case there is no match for a given predicate, we expand that predicate using synsets from WordNet \cite{kilgarriff2000wordnet}.
	(iv) \emph{connecting triple patterns:}
	to generate a federated query, each triple pattern from a parent vertex has to be connected to a triple pattern from the child vertex.
	This connection is via a shared variable, and a $\uparrow$ operator (white circle) is utilized to indicate the positions of the common variable in both vertices.
	In other words, a given parent vertex is connected to its children by a shared variable in either \texttt{object-subject} joint, \texttt{object-object} joint,  \texttt{subject-subject} joint,
	or \texttt{subject-object} joint. 
	After generating queries, we shape a tree of queries. In a hybrid QA, each query is simultaneously executed over each individual source (KG or corpus). In the case of vertices being non-leaves, they are executed after the execution of their children in order to be able to populate their free variables.
	
	%******************************Evaluation****
	\section{Experiments}
	\label{sec:Evaluation}
	
	\subsection{\textbf{Benchmarks}}
	We evaluated HCqa from the various dimensions that we contributd to. Thus, several rounds of the evaluation were conducted on various gold standards.
	We rely on the existing gold standards from the state-of-the-art which are listed as follows:  
	\begin{itemize}
		\item \textbf{Textual corpus}\footnote{\url{http://www.mpi-inf.mpg.de/departments/d5/software/clausie}} is compiled from three datasets (i) New York Times, (ii) Reverb and (iii) Wikipedia.
		
		\item  \textbf{QALD } challenge held annually from 2011. We only consider the task of hybrid in QALD-5\cite{unger20155th} and QALD-6\cite{unger20166th}, where the associated questions are relatively long and complicated and require integrating heterogeneous sources to exploit the final answer.
		\item \textbf{Hybrid Corpus}\cite{grau2018corpus}: is a corpus consists of 4300 pairs of questions and answers, which have been collected from the CLEF and TREC databases and have been enriched with entities and relationships from a knowledge base.
		\item \textbf{ComplexQuestion}\cite{abujabal2017automated} is a benchmark included multiple clauses questions constructed using the crawl of WikiAnswers
		(http://wiki.answers.com). The gold standard answer sets are from Freebase.
	\end{itemize}

	\subsection{Evaluations}
	We evaluated the performance of our approach concerning the two modules which we contributed, i.e., (i) relation extraction module and (ii) Aggregation of answer set module. Finally, the performance of answer extraction for our system was evaluated in two parts of hybrid and complex question answering systems, separately. In the following, we individually present these evaluation scenarios.
	\subsubsection{\textbf{Evaluating relation extraction module}} 
	The majority of the state-of-art of relation extraction approaches are evaluated on a textual corpus, while there is no sufficient attention on the short, informal and noisy text such as user-supplied input query or only the final answer is existed, and generated relations are not accessible\cite{abujabal2017automated}. 
	In this experiment, we employ both types of corpora, i.e., textual corpus and query inventory. We use textual corpus which was also applied by the prior art ClausIE \cite{DelCorro2013} and LS3RYIE\cite{vo2017self}. 
	The second source for our experiment is a query inventory from the hybrid task of the QALD-6 challenge\footnote{\url{https://qald.sebastianwalter.org/index.php?x=challenge&q=6}} \cite{unger20166th}. We compare our module with the recent work,  LS3RyIE, presented in \cite{vo2017self}, as we call it the baseline, hereafter.
	
	\paragraph{\textbf{Relation extraction on textual corpus}.} We run our module on the employed the textual corpus to automatically extract relations from the underlying corpus. To evaluate its effectiveness, we rely on human judge using two annotators being linguistic. 
	They annotated the extracted relations with `true' or `false' respectively indicating the given relation is correct or not.
	We computed the agreement rate between the annotators using Cohen's kappa \cite{cohen1960coefficient} with 0.65 for New York Times dataset, 0.61 for Reverb dataset and 0.70 for the Wikipedia dataset.
	Similar to the art, we use the precision metric ($P=\frac{\#\text{correct relations}}{\#\text{total relations}}$) to evaluate the accuracy.
	Since yet there is no gold standard, calculating recall is not feasible.
	Table \ref{tab:IVCS} shows the ratio of precision for each type of relations in the optimum setting (details of the settings are discussed later).
	We achieved satisfactory precision in most of the cases. We will discuss error analysis later.\\
	\begin{table}[hpt]\caption {The precision ratio (i.e., number of correct relations over the total number of extracted relations) for each type of relationship over the three corpora. GEN. stands for Genitive \& Preposition, APP. stands for Appositive, NOUN. stands for Noun Phrase, V. stands for Verbal, C. OR S. stands for Comparative Or Superlative, ADJ. stands for Possessive Adjective+Whose relation.  } \label{tab:IVCS}
		\vspace{2mm}
		\centering
		\scriptsize
		\begin{center}
			\begin{tabular}{ l|c|c|c|c|c|c} 
				
				\toprule
				\centering
				\textbf{ Corpus}   &\textbf{ Gen.} &
				\textbf{ App.}&
				\textbf{ Noun.}&
				
				\textbf{ C. or S.}&
				\textbf{ V.}&
				\textbf{ Adj.}
				\\
				\midrule
				\multirow{1}{*}{\textbf{NYTimes}} & $262/282$&$26/41$&$344/360$&$4/4$&$605/851$&$52/52$\\ 
				\multirow{1}{*}{\textbf{Reverb}} & $628/668$&$43/65$&$925/944$&$12/12$&$1491/2139$&$78/85$ \\
				\multirow{1}{*}{\textbf{wikipedia}} & $249/268$&$22/34$&$287/290$&$6/6$&$628/828$&$41/48$\\ 
				
				\bottomrule
			\end{tabular}
			
		\end{center}
	\end{table}
	To have a deeper insight on the performance of the relation extraction module, we consider the following settings concerning to our earlier discussion introduced in Section \ref{sec:problemStatement} concerning the three relation types including Genitive \& Preposition, Appositive and Noun Phrase.

	\begin{itemize}
		
		\item{\emph{Minimal noun phrase extraction.}} This setting denoted by A refers to the minimal strategy for detecting noun phrases.
		\item{\emph{Quotation marks and expression consideration.}} This setting denoted by B  considers quotation marks and expressions for detecting noun phrases.
		\item{\emph{Named Entity consideration.}} This setting denoted by C considers named entities.
		
		\item{\emph{Genitive \& preposition reformation.}} This setting denoted by D reforms the Genitive \& Preposition relation.
		
		\item{\emph{Appositive relation filtering.}} 
		This setting denoted by E filters certain Appositive relations (according to the conditions discussed earlier).
		
		\item{\emph{Relation extraction with all argument types:}} This setting denoted by F  extends the approach \cite{vo2017self} by including any possible argument (e.g., object, subject, and complement in the dependency parse tree). 
		
		\item{\emph{Noun phrase extraction I.}} This setting denoted by G considers approach I, for noun phrase relation extraction.
		
		\item{\emph{Noun phrase extraction II.}}  This setting denoted by H considers approach I for noun phrase relation extraction.
	\end{itemize}
	Furthermore, we consider settings which are a composition of the above settings as follows:
	
	\begin{itemize}
		\item{Genitive \& Preposition relation settings:} A, AB, ABC, ABD, ABCD, ABCDF 
		\item{Appositive relation settings: } A, AB, ABC,ABE, ABCE, ABCEF 
		\item{Noun Phrase relation settings: }
		G, B, C, H, BCH 
	\end{itemize}
	It should be noted that we do not define any setting for the following three types of relations including Verbal, Possessive Adjective + Whose and Comparative or superlative. Because, e.g., the precision of Comparative or Superlative relations was flawless (see the precision in Table \ref{tab:IVCS}).
	In the following, we present our detailed discussion upon the evaluation of the designed settings.
	Our baseline considers only two relation types including Genitive \& Preposition and Appositive relations.
	In Table \ref{SentencesExample}, we provide samples of sentences containing the concerning relations.
	There, also we represent the relations extracted via our module versus the baseline.
	
	\begin{table}[hpt]
		\centering
		\scriptsize
		\caption{Sentence examples for extracted relation and their labels based on applied setting and relation type for HCqa(our system) and LS3RyIE(baseline).}
		\vspace{2mm}
		\label{SentencesExample}
		\begin{tabular}{|c|l|p{0.6cm}|p{6.4cm}|c|}
			\toprule
			\textbf{Setting} &
			\textbf{Relation} & \multicolumn{1}{c|}{\textbf{System}} & \multicolumn{1}{c|}{\textbf{Setting-related extracted relation (s, p, o)}} &
			\textbf{tag} \\ \midrule

			\multicolumn{5}{|l|}{\begin{tabular}[c]{@{}l@{}}\textbf{S1:} "Martin Gibson is the company 's chairman and has served as a director of the\\ parent company since 1992.”\end{tabular}} \\ \hline
			\multirow{3}{0.1cm}{\begin{tabular}[c]{@{}c@{}}\textbf{A}\end{tabular}}& \multirow{3}{1.2cm}{\begin{tabular}[c]{@{}c@{}}{Genitive \&}\\{Preposition}\end{tabular}} &\multirow{1}{0.3cm} HCqa & company, has, chairman & true \\ \cline{3-5} 
			&  & baseline & \begin{tabular}[c]{@{}l@{}}company, has, chairman and has served as a director of\\ the parent company since 1992\end{tabular} & false \\ \toprule

			\multicolumn{5}{|l|}{\begin{tabular}[c]{@{}l@{}}\textbf{S2:} "Doctors in Pennsylvania and West Virginia are expected to notify S.M.I. bioterror \\experts of any "suspicious event," from an unusual rash to a finger lost in an explosion,\\ identifying but not informing the patient”\end{tabular}}\\ \hline
			\multirow{3}{0.1cm}{\begin{tabular}[c]{@{}c@{}}\textbf{A}\end{tabular}}& \multirow{3}{1.2cm}{\begin{tabular}[c]{@{}c@{}}{Genitive \&}\\{Preposition}\end{tabular}} & \multirow{2}{*}{HCqa} & doctors, are in , Pennsylvania & true\\&&&doctors, are in, West Virginia & true\\\cline{3-5}&& baseline &\begin{tabular}[c]{@{}l@{}}doctors, are in, Pennsylvania and west Virginia\end{tabular} & true \\ \toprule

			\multicolumn{5}{|l|}{\textbf{S3:} "In March 2009, Ludwig appeared in a lead role in Disney's "Race to Witch Mountain.”} \\ \hline
			\multirow{2}{0.1cm}{\begin{tabular}[c]{@{}c@{}}\textbf{B}\end{tabular}}& \multirow{2}{1.2cm}{\begin{tabular}[c]{@{}c@{}}{Genitive \&}\\{Preposition}\end{tabular}} & HCqa & Disney, has, Race to Witch Mountain & true \\ \cline{3-5} 
			& &baseline & \begin{tabular}[c]{@{}l@{}}Disney, has, Race\end{tabular} & false \\ \toprule

			\multicolumn{5}{|l|}{\begin{tabular}[c]{@{}l@{}}\textbf{S4:} "The menu, imported from Ben's Deli in Queens, includes matzoh ball soup,\\ corned beef and pastrami sandwiches, chopped liver, kishke and knishes.”\\ "Ben's Deli" is a Named Entity\end{tabular}}\\ \hline
			\multirow{2}{0.1cm}{\begin{tabular}[c]{@{}c@{}}\textbf{C}\end{tabular}}& \multirow{2}{1.2cm}{\begin{tabular}[c]{@{}c@{}}{Genitive \&}\\{Preposition}\end{tabular}} & HCqa & -& - \\ \cline{3-5} 
			& &baseline & \begin{tabular}[c]{@{}l@{}}Ben, has, Deli in Queens\end{tabular} & false \\ \toprule

			\multicolumn{5}{|l|}{\textbf{S5:} "The statue stands 56 feet tall and was placed atop Red Mountain in 1936.”} \\ \hline
			\multirow{2}{0.1cm}{\begin{tabular}[c]{@{}c@{}}\textbf{D}\end{tabular}}& \multirow{2}{1.2cm}{\begin{tabular}[c]{@{}c@{}}{Genitive \&}\\{Preposition}\end{tabular}} & HCqa & - & - \\ \cline{3-5} 
			&& baseline & \begin{tabular}[c]{@{}l@{}}red mountain, is in, 1936\end{tabular} & false \\ \toprule

			\multicolumn{5}{|l|}{\begin{tabular}[c]{@{}l@{}}\textbf{S6:} "Terrorist attacks by E.T.A. have declined in recent years and the number of its hardcore\\ militants is thought to have fallen from the hundreds of 15 years ago to several score.”\end{tabular}}\\ \toprule
			\multirow{2}{0.1cm}{\begin{tabular}[c]{@{}c@{}}\textbf{F}\end{tabular}}& \multirow{2}{1.2cm}{\begin{tabular}[c]{@{}c@{}}{Genitive \&}\\{Preposition}\end{tabular}} & HCqa & Terrorist attacks, is by , E.T.A & true \\ \cline{3-5} 
			& &baseline & \begin{tabular}[c]{@{}l@{}}-\end{tabular} & - \\ \toprule

			\multicolumn{5}{|l|}{\begin{tabular}[c]{@{}l@{}}\textbf{S7:} "More widely, in late August, 2007 the group was accused in "The Telegraph", a\\ conservative British newspaper, of torturing, detaining, and firing on\\ unarmed protesters who had objected to policies of the Hamas government.”\end{tabular}}\\ \hline
			\multirow{2}{0.1cm}{\begin{tabular}[c]{@{}c@{}}\textbf{A}\end{tabular}}& \multirow{2}{1.2cm}{Appositive} & \multirow{1}{0.5cm}{HCqa} &The Telegraph, is, a conservative British newspaper & true\\\cline{3-5}&&  \multirow{1}{0.3cm}{baseline} & \begin{tabular}[c]{@{}l@{}}The Telegraph of torturing detaining and firing on unarmed\\ protesters, is, a conservative British newspaper\end{tabular}& false\\ \toprule

			\multicolumn{5}{|l|}{\begin{tabular}[c]{@{}l@{}}\textbf{S8:} "During the Punitive Expedition into Mexico in 1916, he was for a time depot manager\\ at Columbus, New Mexico, the main logistical base of the expedition. \\"Columbus, New Mexico" is a Named Entity\end{tabular}}\\ \hline
			\multirow{2}{0.1cm}{\begin{tabular}[c]{@{}c@{}}\textbf{C}\end{tabular}}&
			\multirow{2}{1.2cm}{Appositive} & HCqa & - & - \\ \cline{3-5} 
			&& baseline & \begin{tabular}[c]{@{}l@{}}Columbus, is, New Mexico\end{tabular} & false \\ \toprule

			\multicolumn{5}{|l|}{\begin{tabular}[c]{@{}l@{}}\textbf{S9:} "In its most recent survey, the Congress for New Urbanism, a nonprofit organization\\ based in Chicago, reported 648 neighborhood-scale New Urbanist communities in the\\ United States, an increase of 176 over a 12-month period."\end{tabular}}\\ \hline
			\multirow{2}{0.1cm}{\begin{tabular}[c]{@{}c@{}}\textbf{E}\end{tabular}} &\multirow{2}{1.2cm}{Appositive}& HCqa & - & - \\ \cline{3-5} 
			& & \multirow{1}{0.3cm}{baseline} &the United States, is, an increase of 176 & false\\ \toprule

			\multicolumn{5}{|l|}{\begin{tabular}[c]{@{}l@{}}\textbf{S10:} "In the most world-renowned episode , seven frantic photographers on motorbikes\\ chased Princess Diana and\\ her companion , Dodi al-Fayed , after leaving the Ritz hotel in Paris.”\end{tabular}}\\ \hline
			\multirow{2}{0.1cm}{\begin{tabular}[c]{@{}c@{}}\textbf{F}\end{tabular}} &\multirow{2}{1.2cm}{Appositive}& HCqa & her companion, is, Dodi al-Fayed & true \\ \cline{3-5} 
			& & \multirow{1}{0.3cm}{baseline} &- & -\\ \toprule

		\end{tabular}
	\end{table}
	
\begin{figure}[ht]
		\centering
		\includegraphics[width =0.5\textwidth]{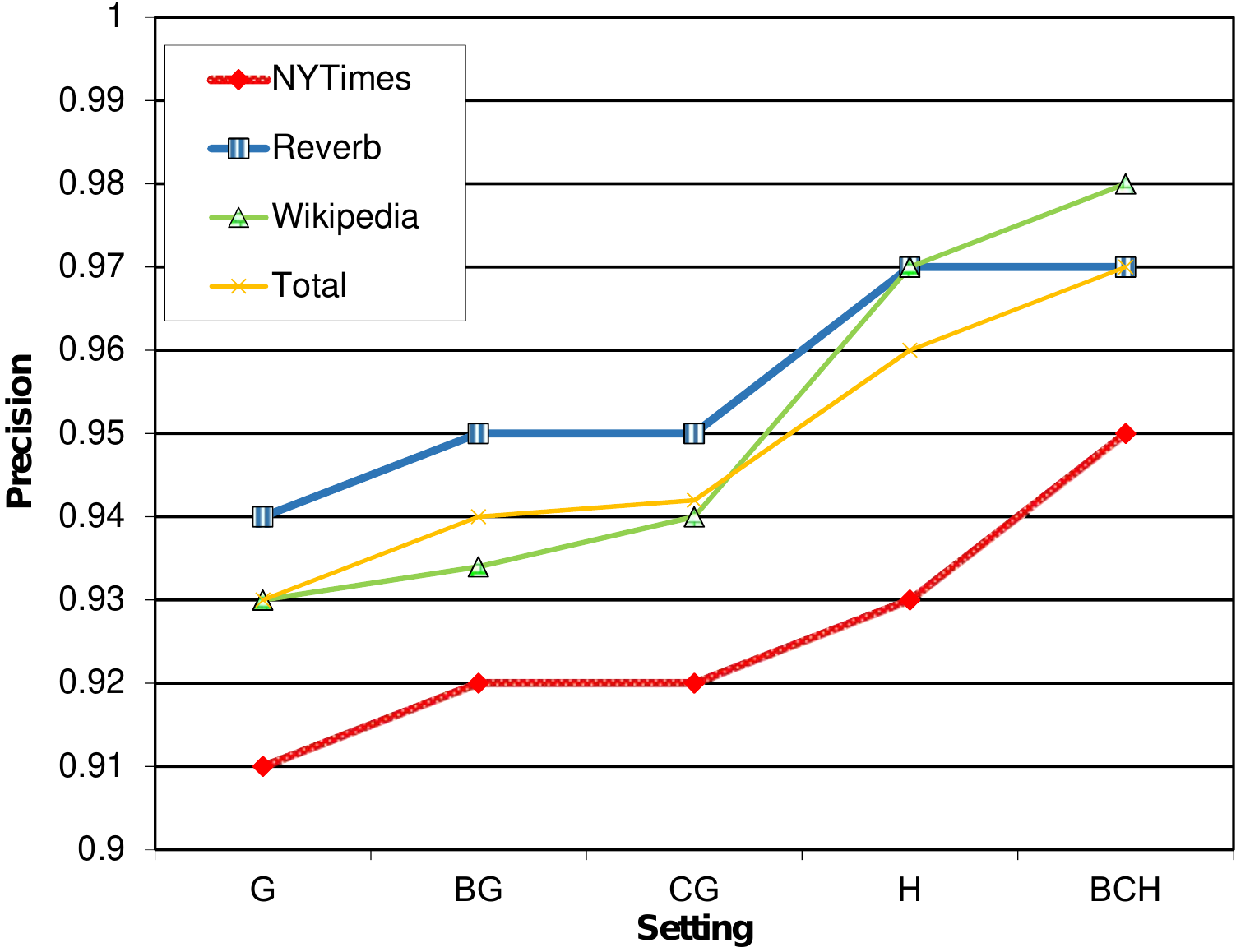}
		\caption{The precision ratio of Noun Phrase relations in different benchmarks with different settings.}
		\label{fig:NounPhrase}
	\end{figure}

	\begin{itemize}
		\item \emph{Genitive \& Preposition relations evaluation:}
		Fig. \ref{fig:GenN.pdf} represents the precision of this type of relation and 
		Fig. \ref{fig:GenNN.pdf} represents the number of extracted relations.
		Concerning setting A, two bold conclusions are observed: 1) there is an increase in precision in comparison to the baseline, e.g., the sentence S1 in Table \ref{SentencesExample} shows the success of our approach in detecting this kind of relation.  2) there is a trivial increase in the number of extracted relations compared to the baseline, e.g., the sentence S2 in Table \ref{SentencesExample} illustrates our success in detecting this type of relation.
		As we can notice from Fig. \ref{fig:GenN.pdf}, the settings B, C, D, and F reached an increase in precision metric comparing to the baseline. 
		The sentences S3, S4, S5 and S6 sentences in Table \ref{SentencesExample} exemplify these settings. 
		Furthermore, we run the composition settings  ABC and ABD. The results of this experiment (Fig. \ref{fig:GenN.pdf}) shows that ABC outperforms ABD pointing out the importance of ``Named Entity consideration'' rather than ``Genitive \& Preposition reformation''. 
		Finally, adding the setting F led to a considerable decrease in precision while it increases the total number of extracted relations.
		%******************genitive \& chart**********
		
		\begin{figure}[htbp]
			\begin{center}
				\subfloat[]{\includegraphics[width =0.5\textwidth]{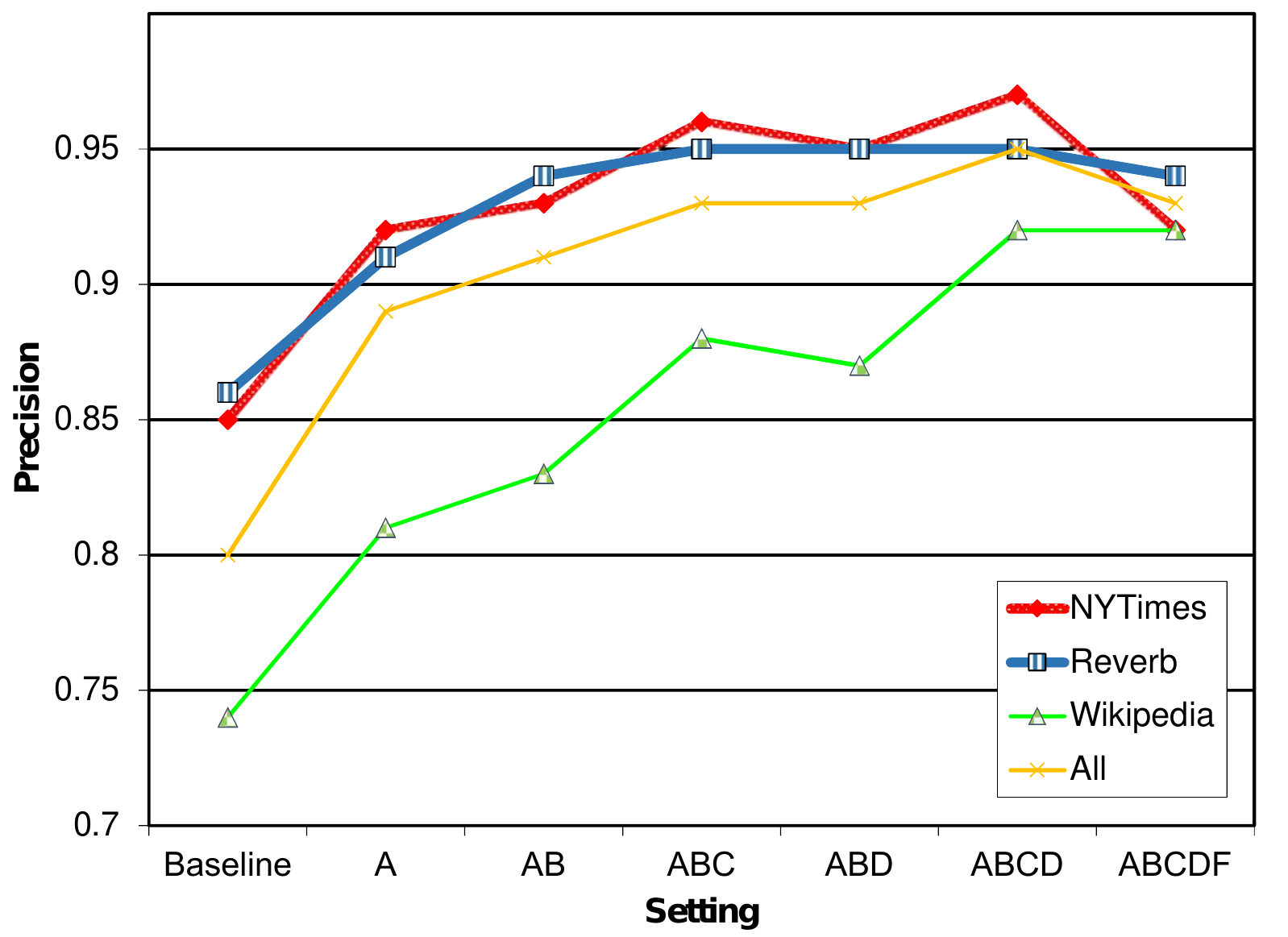}\label{fig:GenN.pdf}} 
				\subfloat[]{\includegraphics[width =0.5\textwidth]{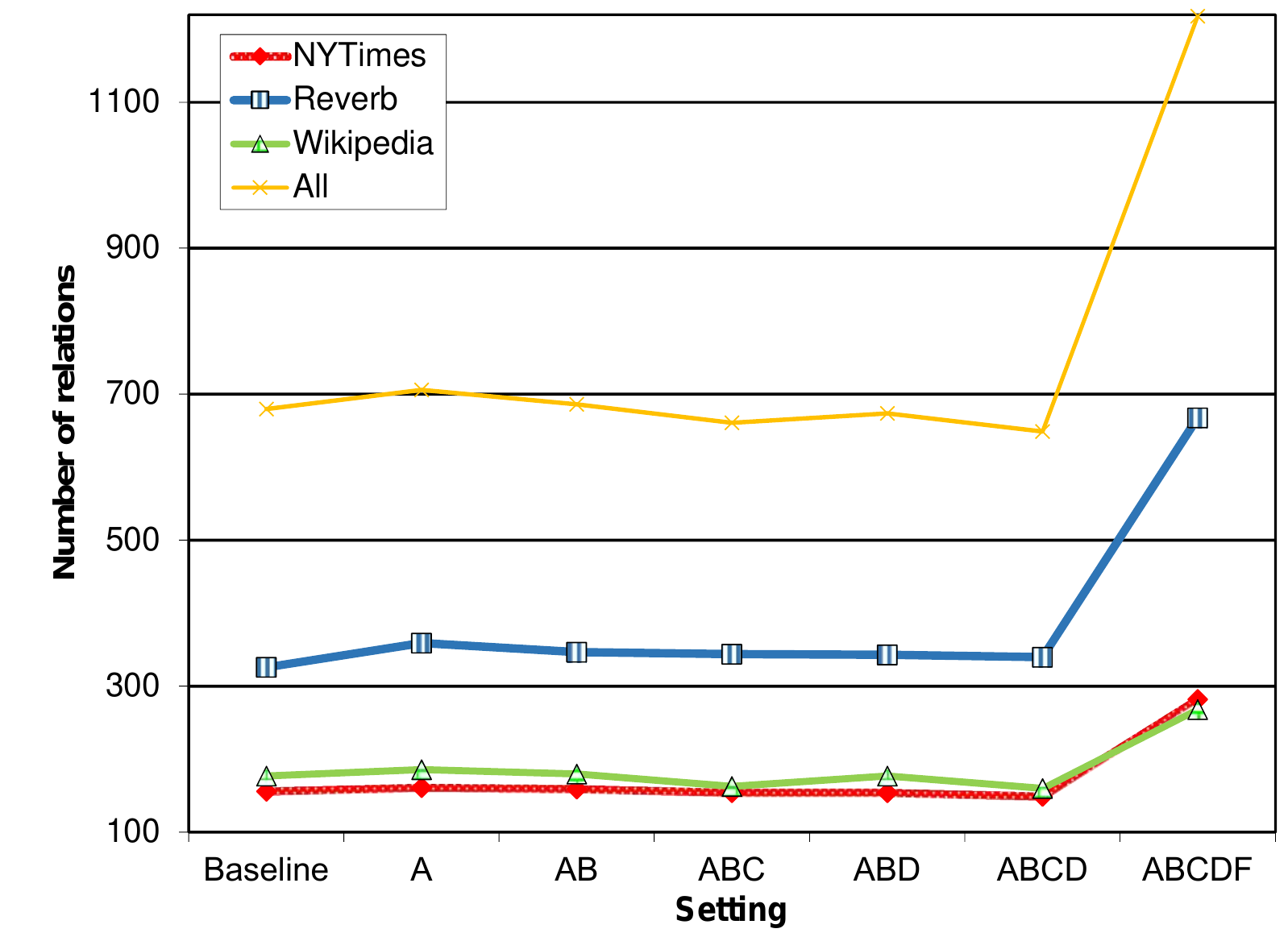}\label{fig:GenNN.pdf}} \\
				
				\caption{(a) The precision ratio and (b) the number of Genitive \& Preposition relations for different settings.}
				\label{fig:genitive}
			\end{center} 
		\end{figure}
		
		\item \emph{Appositive relation:}  Fig. \ref{fig:preComma} and Fig. \ref{fig:NumComma} show the performance of the settings related to this type of relation. Using the settings A, C, E, and F significantly increase the precision metric rather comparing to the baseline. 
		Concerning the number of relations (recall), the settings A, C, and E resulted in a lower recall, and higher precision whereas adding the setting F leads to higher recall and lower precision.
		Comparing the composition settings ABC and ABE, shows the higher precision in ABC rather than ABD meaning Named Entity is more effective than Appositive Relation Filtering.
		The sampled sentences S7, S8, S9 and S10 in Table \ref{SentencesExample} shows the success and failure cases of our settings compared to the baseline.

		%***************Comma chart*************** 
		
		\begin{figure}[htbp]
			\begin{center}
				\subfloat[]{\includegraphics[width =0.5\textwidth]{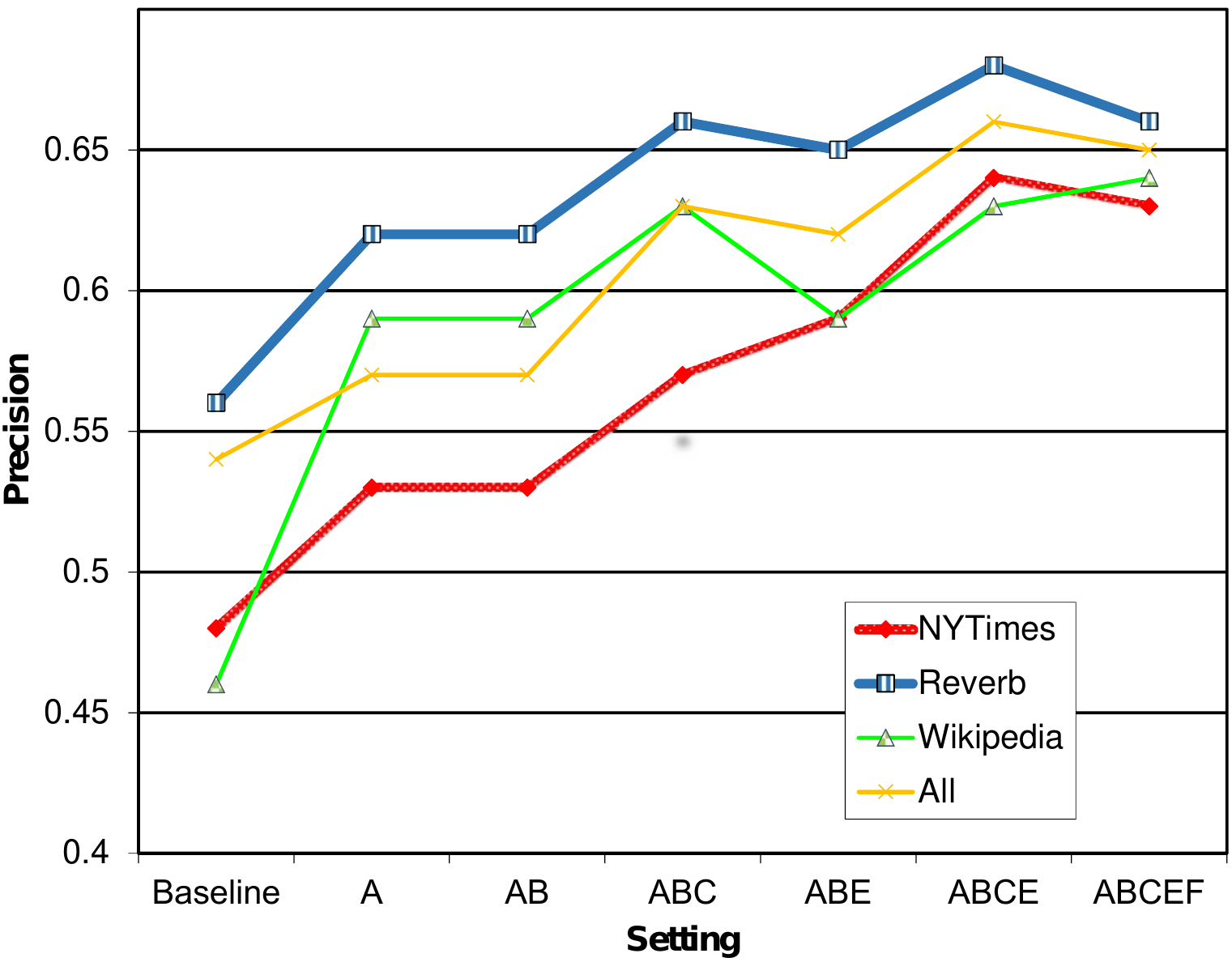}\label{fig:preComma}} 
				\subfloat[]{\includegraphics[width =0.5\textwidth]{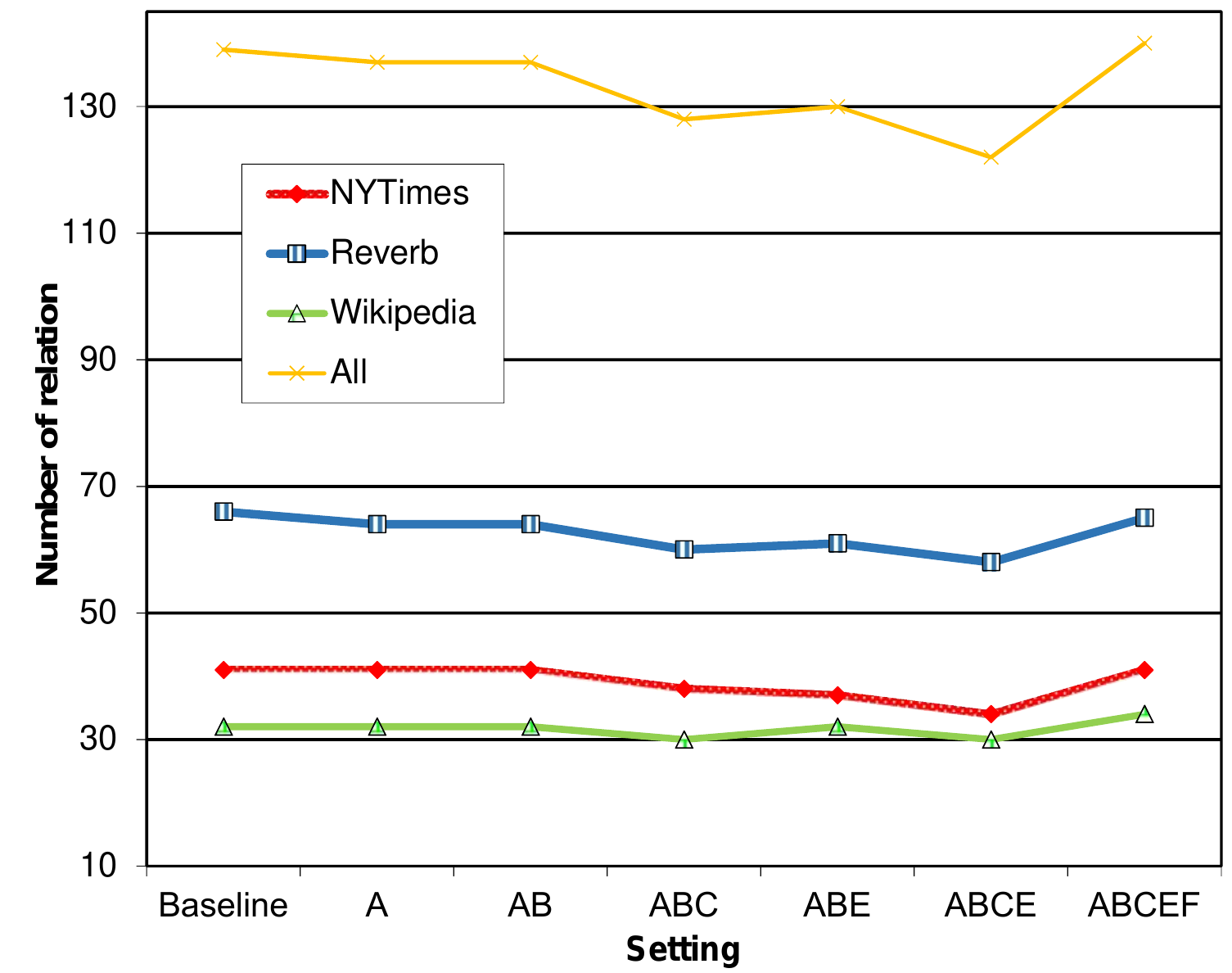}\label{fig:NumComma}} \\
				
				\caption{(a) The precision ratio and (b) the number of Appositive relations over various settings.}
				\label{fig:Comma}
				j	\end{center} 
		\end{figure}
		\item \emph{Noun Phrase relation:}
		We applied the two settings H and G concerning the approach I and II concerning Noun Phrase relation extraction.
		Fig. \ref{fig:NounPhrase} represents the precision on the various corpora, i.e., NYTimes, Reverb, Wikipedia and the compiled corpus (Total) using the settings  B, C along with H and G.
		The results show that the setting C is trivially effective than B and the setting H is superior to G.

	\end{itemize}
	We compared our proposed approach with the results of OIE systems reported in \cite{vo2017self} over precision metric in Table \ref{tab:OpenIESystems}. The total precision for three corpus in our system is $81/74\%$ while the state-of-art system, LS3RyIE results $68/33\%$ for precision.

	\begin{table}[hpt]\caption {The precision ratio (i.e., number of correct relations over the total number of extracted relations) results of OIE systems for each free text benchmark.  } \label{tab:OpenIESystems}
		\vspace{2mm}
		\centering
		\scriptsize
		\begin{center}
			\begin{tabular}{ l|c|c|c} 
				
				\toprule
				\centering
				\textbf{ System}   &\textbf{ ReVerb} &
				\textbf{ Wikipedia}&
				\textbf{ NYT}
				\\
				\midrule
				\multirow{1}{*}{\textbf{TextRunner}} & $35/84\%(286/798)$&$-$&$-$\\ 
				\multirow{1}{*}{\textbf{WOE}} & $43/48\%(447/1028)$&$-$&$-$\\ 
				\multirow{1}{*}{\textbf{ReVerb}} & $53/37\%(388/727)$&$66/26\%(165/249)$&$54/98\%(149/271)$\\  
				\multirow{1}{*}{\textbf{OLLIE}} & $44/04\%(547/1242)$&$41/41\%(234/565)$&$42/46\%(211/497)$\\
				\multirow{1}{*}{\textbf{ClausIE}} & $50/37\%(1182/2348)$&$49/56\%(397/797)$&$52/67\%(493/936)$\\
				\multirow{1}{*}{\textbf{LS3RyIE}} & $67/77\%(1642/2425)$&$68\%(614/903)$&$70/19\%(690/983)$\\
				\multirow{1}{*}{\textbf{HCqa}} & $\textbf{81/19\%(3177/3913)}$&$\textbf{83/64\%(1233/1474)}$&$\textbf{81/32\%(1293/1590)}$\\
				\bottomrule
			\end{tabular}
			
		\end{center}
	\end{table}
	
	\paragraph{\textbf{Error analysis:}}
	To obtain further insight on the performance of relation extraction module, particularly for ``Noun Phrase relation extraction'', we provide the error analysis as follows:
	\begin{itemize}
		\item{\emph{Quotation marks and expression consideration:}} If ``quotation marks and expressions'' is not applied, then wrong relations might be inferred, e.g., for the given sentence \texttt{`He graduated summa cum laude from the University of California, Los Angeles.'}, \texttt{`summa'} is an adjective phrase and \texttt{`cum'} and \texttt{`laude'} are two noun phrases. The approach II yields in inferring the two separate noun phrases: (i) \texttt{`summa laude'} and \texttt{`cum laude'}, whereas the desired relation is \texttt{`summa cum laude'} is an expression.

		\item{\emph{Named Entity:}} 
		If ``Named Entity'' is not considered, it also yields in inferring wrong relations, e.g., \texttt{`North Korean forces'} contains \texttt{`North'} and \texttt{`Korean'} as adjective phrases and \texttt{`forces'} as a noun phrase. Without  considering Named Entity, the two extracted pairs are: (i) \texttt{`North forces'} and (ii) \texttt{`Korean forces'}, while \texttt{`North Korean"} is itself a Named Entity.
		
		\item{\emph{Noun phrase extraction using the approach I:}} 
		This approach pairs the nearby words which might lead to extracting wrong relations, e.g., in the given noun phrase \texttt{`English crime writer'}, the two paired words are (i) \texttt{`English crime'} and (ii) \texttt{`crime writer'} where the first one is faulty.  
		
		\item{\emph{POS tagging:}} Since we applied an external tool for POS tagging, errors of this tool affect our performance, e.g., for the given sentence \texttt{`I have also learned that lice plan their arrival to maximize \\household stress.'}, the words \texttt{`lice'} and \texttt{`plan'} are respectively tagged with \texttt{`NN'} and \texttt{`NN'} whereas \texttt{`plan'} is a verb. This fault leads to inferring the wrong paired word \texttt{lice plan}.
		
		\item{\emph{Independent-Meaning:}} This case includes two situations as follows: (i) addresses: our module breaks the addresses such as the noun phrase \texttt{`165 West 65th Street'} into smaller segments \texttt{`165'},\texttt{`West'}, \texttt{`65th'} and \texttt{`Street'}.
		(ii) noun phrases which modify the verb, e.g, in the sentence \texttt{`Two independent journalists went on trial today'}, \texttt{`today'} is a time modifier, and is independent of \texttt{`trial'} while our module considers them together.

	\end{itemize}
	
	%     
	%********************Noun Phrase chart and table
	
	%\end{itemize}
	
	\begin{comment}
	
	The numbers for relations caused by different errors are jrepresented in Table \ref{tab:NpError}.   
	\begin{table}[hpt]\caption {Number of failures in extracting NounPhrase relation. }
	\vspace{2mm}
	\label{tab:NpError}
	\begin{center}
	\scriptsize
	\begin{tabular}{ |l|c|c|c|c|c|c|c| } 
	\hline
	& \textbf{Tag} &
	\textbf{expression}&
	\textbf{Parser}&
	\textbf{Meaning}&
	\textbf{NER}&
	\textbf{approach I}
	
	\\
	\hline
	\multirow{1}{*}{NYTimes dataset(375)} & $3$&4 &8 &6 &2 &9\\ 
	\multirow{1}{*}{Reverb dataset(975)} & $1$&1 &12 & 6&2 &28 \\
	\multirow{1}{*}{wikipedia dataset(306)} & $0$& 0& 3&0 &2 &15 \\
	
	\hline
	\end{tabular}
	\end{center}
	\end{table}
	\end{comment}
	
	\begin{figure}[htbp]
		
		\begin{center}
			\subfloat[]{\includegraphics[width =1\textwidth]{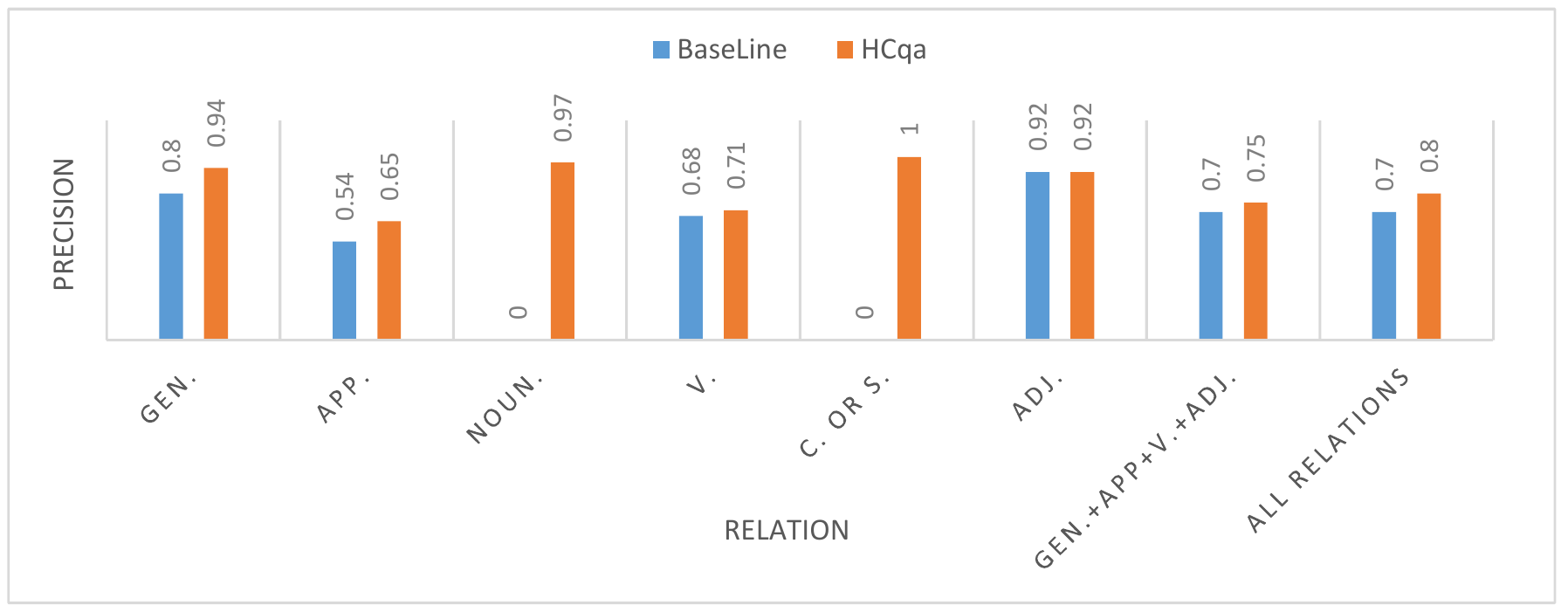}\label{fig:allRelCmp.PNG}} \\
			\subfloat[]{\includegraphics[width =1\textwidth]{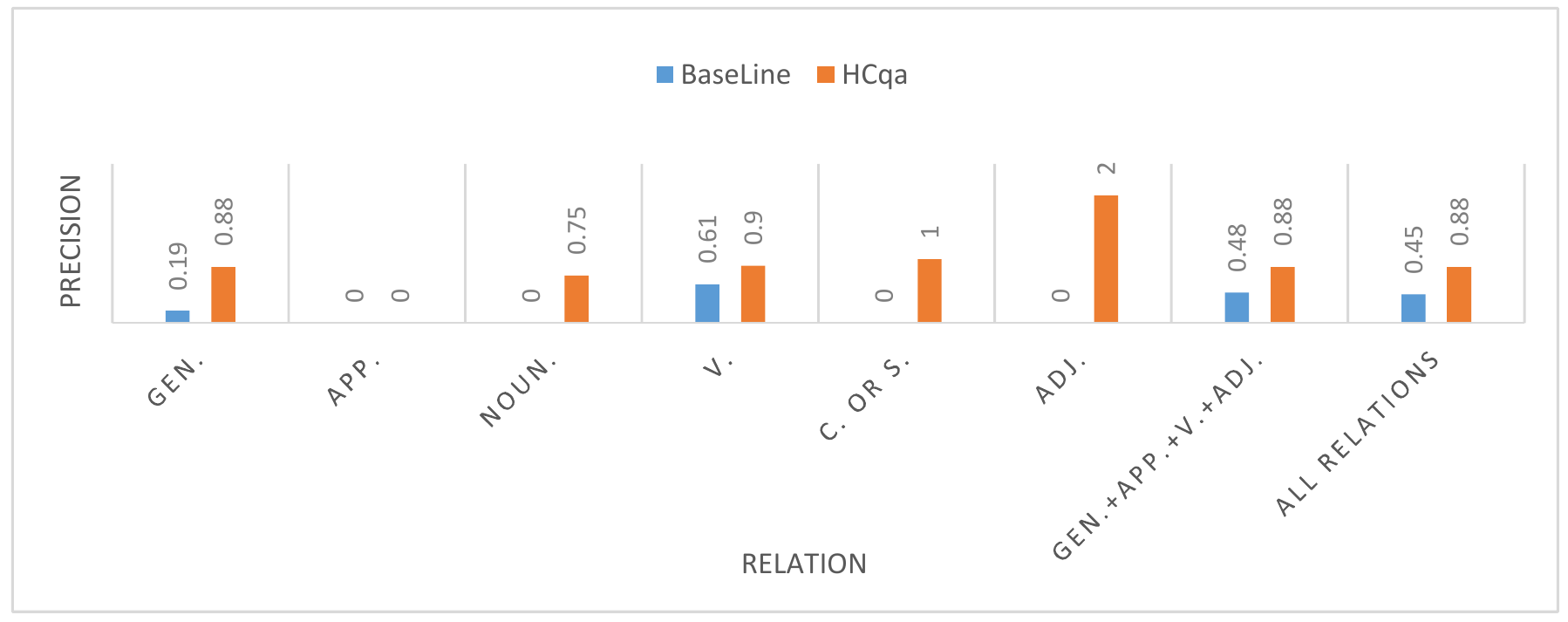}\label{fig:allRelCmpQ.PNG}} \\
			
			\caption{The comparison between our approach and baseline for (a) textual corpus and  (b) query inventory for all extracted relations. GEN. stands for Genitive \& Preposition, APP. stands for Appositive, NOUN. stands for Noun Phrase, V. stands for Verbal, C. OR S. stands for Comparative Or Superlative, ADJ. stands for Possessive Adjective+Whose relation. }
			\label{fig:Cmp}
		\end{center} 
	\end{figure}
	\begin{table}[hpt]\caption {The precision ratio (i.e., number of correct relations over the total number of extracted relations) for each type of relationship over the three corpora. GEN. stands for Genitive \& Preposition, APP. stands for Appositive, NOUN. stands for Noun Phrase, V. stands for Verbal, C. OR S. stands for Comparative Or Superlative, ADJ. stands for Possessive Adjective+Whose relation.  } \label{tab:IVCS}
		\vspace{2mm}
		\centering
		\scriptsize
		\begin{center}
			\begin{tabular}{ l|c|c|c|c|c|c} 
				
				\hline
				\centering
				\textbf{ Corpus}   &\textbf{ Gen.} &
				\textbf{ App.}&
				\textbf{ Noun.}&
				
				\textbf{ C. or S.}&
				\textbf{ V.}&
				\textbf{ Adj.}
				\\
				\hline
				\multirow{1}{*}{\textbf{NYTimes}} & $262/282$&$26/41$&$344/360$&$4/4$&$605/851$&$52/52$\\ 
				\multirow{1}{*}{\textbf{Reverb}} & $628/668$&$43/65$&$925/944$&$12/12$&$1491/2139$&$78/85$ \\
				\multirow{1}{*}{\textbf{wikipedia}} & $249/268$&$22/34$&$287/290$&$6/6$&$628/828$&$41/48$\\ 
				
				\hline
			\end{tabular}
			
		\end{center}
	\end{table}
	
\begin{figure}[ht]
		\begin{center}
			\subfloat[Formal query]{\includegraphics[width =0.49\textwidth]{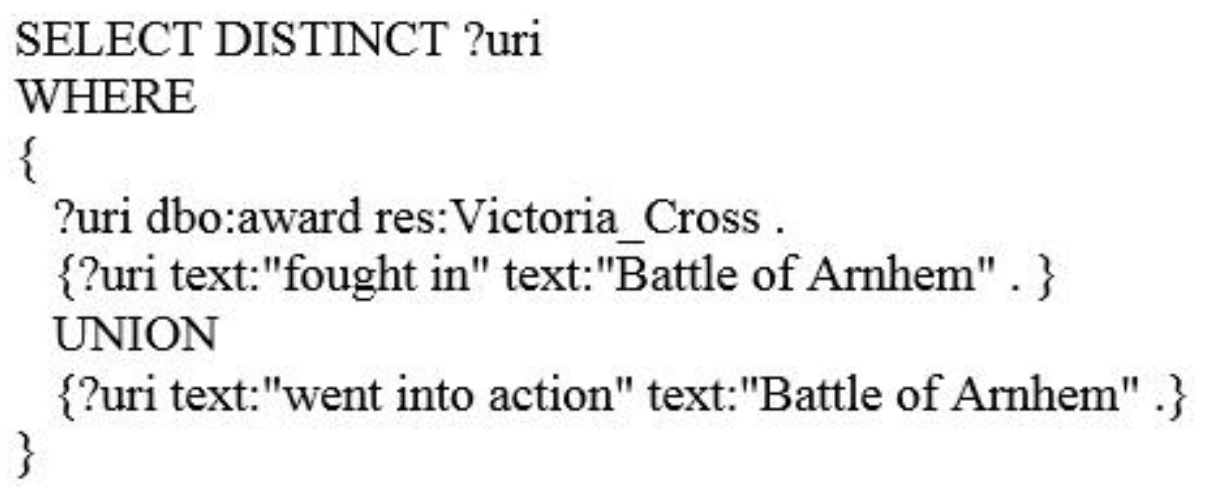}\label{fig:uri.pdf}} 
			\subfloat[Textual Serialization]{\includegraphics[width =0.49\textwidth]{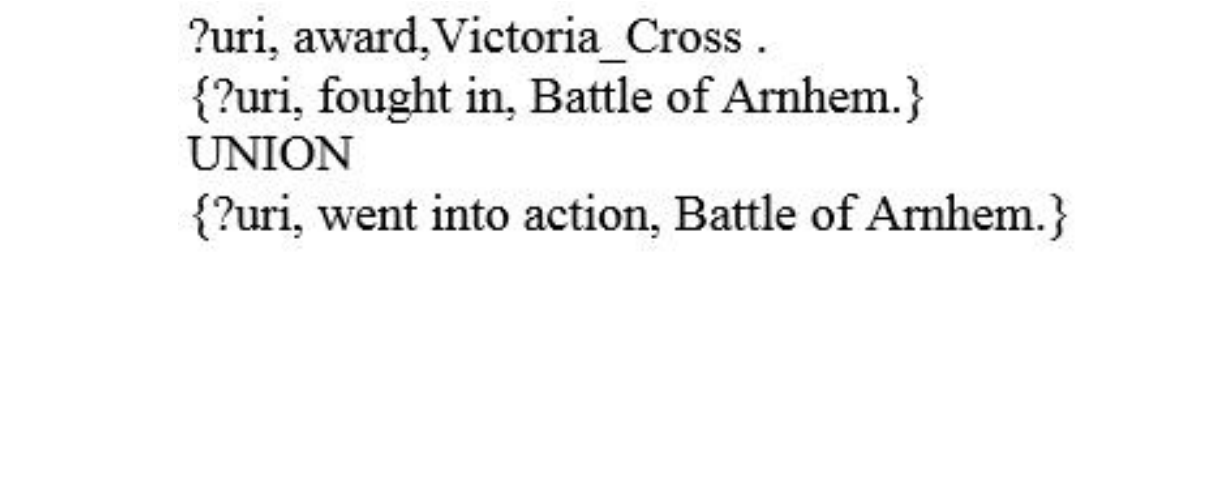}\label{fig:triples.pdf}} \\
			
			\caption{A sample of a formal query from our query inventory for the running example Q4.}
			\label{fig:psudo}
		\end{center} 
	\end{figure}

	\paragraph{\textbf{Relation extraction on query inventory.}} The questions in hybrid task of QALD-6 challenge\footnote{\url{https://qald.sebastianwalter.org/index.php?x=challenge&q=6}} \cite{unger20166th}, containing 75 questions, are employed for our evaluation.
	These questions are annotated with
	the corresponding answers as well as the equivalent formal query (triple patterns).
	We transform all the formal queries to semi-textual serialization queries (dropping URIs).
	Regarding the running example question Q4, its formal query and the transformed representation are shown in Figure \ref{fig:uri.pdf} and \ref{fig:triples.pdf}. 
	Initially, we run Stanford parser \cite{Schuster2016} to extract relations in the form of triple patterns (subject, predicate, object) from questions. We furthermore use  WordNet \cite{kilgarriff2000wordnet} to expand predicates of triple patterns (to resolve vocabulary mismatch to some extent).
	We compared the extracted triples patterns with the relations of formal queries in the gold standard. 
	Our approach recognizes $84/04\%$ ; while our baseline recognizes $44/68\%$ of all triple patterns from different categories.
	
	\begin{comment}
	\begin{table}[hpt]\caption {Accuracy (recall, precision and correctness of relation extraction module on queries from QALD-6\cite{unger20166th} benchmark. } \label{tab:RPCCmp} 
	\scriptsize

	\begin{center}
	\scriptsize
	\begin{tabular}{ |l|c|c|c| } 
	\hline
	\textbf{} & \textbf{Recall} 
	& \textbf{Precision}
	& \textbf{correctness}
	\\
	\hline
	\multirow{1}{*}{baseline} & 0.52&0.56&0.73\\ 
	\multirow{1}{*}{Our approach} & 0.92&0.59&0.94\\
	
	\hline
	\end{tabular}
	\end{center}
	\end{table}
	
	\end{comment}
	
	\begin{table}[hpt]\caption {The number of extracted relations of each category for questions used in hybrid task of QALD-6\cite{unger20166th}.} 
		\vspace{2mm}
		\label{tab:triplePercent} 
		\scriptsize
		
		\begin{center}
			\begin{tabular}{ l|c|c } 
				\toprule
				\textbf{Relations} & \textbf{Our approach}&  \textbf{baseline\cite{vo2017self}}\\
				\midrule
				\multirow{1}{*}{Verbal (62)}&51&38\\
				\multirow{1}{*}{Possesive Adjective+Whose(3)} &3 &0\\
				\multirow{1}{*}{Genitive \& Preposition (24)} &21 &4\\
				\multirow{1}{*}{Appositive (0)} &0 &0\\
				\multirow{1}{*}{Noun Phrase (4)} &3&0\\ 
				\multirow{1}{*}{Comparative or Superlative (1)} & 1&0\\
				\bottomrule
			\end{tabular}
		\end{center}
	\end{table}

	\subsubsection{\textbf{Evaluating the aggregation  of answer set}} 
	Among systems targeting hybrid composite questions, only ISOFT deals with the aggregation approach. 
	ISOFT\cite{park2015isoft} is one of the participating systems in the hybrid task of QALD-5\cite{unger20155th} challenge containing 60 questions. Thus, these questions were also used to evaluate our approach.
	We generated the composite question tree and final query, respectively as a result of our system and ISOFT system, for each question. The evaluation relies on two annotators for judging on the generated composite question tree for our system and the generated query of ISOFT, for each given composite question. In case the output is of a system, is labeled correct of the two annotators, then we consider that as a correct output. The agreement rate (Cohen’s kappa) was with $0.80\%$. The precision result for the ISOFT system is $70.82\%$, and for our proposed approach is $92.04\%$.  
	
	\begin{comment}
	
	\begin{table}[hpt]\caption {Comparing execution order in our system and two systems ISOFT and HAWK.} \label{tab:orderComparision}

	\begin{center}
	\begin{tabular}{ |l|c|c|c| } 
	\hline
	&\textbf{ISOFT} & \textbf{HAWK} & \textbf{Proposed approach}\\
	\hline
	\multirow{1}{*}{test5} & $70\%$ & $90\%$ & $100\%$ \\ 
	\hline
	\multirow{1}{*}{train6} & $64\%$&$85\%$& $93\%$\\
	
	\hline
	\end{tabular}
	\end{center}
	\end{table}
	\end{comment}
	\medskip
	
	\subsubsection{Evaluating the performance of answer extraction}
	
	We compare the performance of our approach in extracting answer against three types of systems or corpora: (i) QA systems taken part in the hybrid task of QALD challenge, (ii) QA systems consider complex questions and (iii) the recently published hybrid corpus \cite{grau2018corpus}. In the first step, we compare our results with the latest participating systems in the hybrid task of QALD challenge.
	The results against the QALD-5 \cite{unger20155th} presented in Table \ref{tab:AnswerExtraction} and for QALD-6 \cite{unger20166th} in Table \ref{tab:AnswerExtraction1}. The accuracy metrics are the number of questions which can be processed, recall $\frac{\#\text{correct system answers}}{\#\text{gold standard answers}}$, precision $\frac{\#\text{correct system answers}}{\#\text{system answers}}$, F-1 measure ($\frac{2\ast\text{precision}\ast\text{recall}}{\text{precision}+\text{recall}}$
	) and F-1 global (for all questions) metrics. Xu et.al\cite{feng2016hybrid} also reported the performance of their hybrid question answering system based on only F-1 global metric over QALD-6 benchmark. Results show our system can beat all systems evaluated over the hybrid tasks of QALD-5 and QALD-6 challenges.  
	\begin{table}[hpt]
		\scriptsize
		\begin{center}
			\begin{tabular}{ l|c|c|c|c|c } 
				
				\toprule
				
				\textbf{System}	&\textbf{Processed} &  \textbf{Recall} & \textbf{Precision} & \textbf{F-1} & \textbf{F-1 Global} \\
				\midrule
				\multirow{1}{*}{\textbf{HCqa} (our approach)} & $10$ & $1.00$&$0.7$&$0.81$&$\textbf{0.56}$\\ \multirow{1}{*}{\textbf{ISOFT}} & $3$ &$1.00$&$0.78$&$0.87$&$0.26$\\
				\multirow{1}{*}{\textbf{HAWK}} & $3$ & $0.33$&$0.33$&$0.33$&$0.10$\\
				\multirow{1}{*}{\textbf{YodaQA}} & $10$ &$0.10$&$0.10$&$0.10$&$0.10$\\
				\multirow{1}{*}{\textbf{SemGraphQA}} & $6$ & $0.00$&$0.20$&$0.00$&$0.00$\\
				\multirow{1}{*}{\textbf{Xser}} & $3$ & $0.00$&$0.00$&$0.00$&$0.00$\\
				\bottomrule
			\end{tabular}
		\end{center}
		\caption {Answer extraction comparison with participating systems in hybrid task of QALD-5\cite{unger20155th} challenge over 10 test questions.} \label{tab:AnswerExtraction}  
	\end{table}

	\begin{table}[hpt]\caption {Answer extraction comparison against systems evaluated over QALD-6\cite{unger20166th} challenge over 25 test questions.} \label{tab:AnswerExtraction1}  
		\scriptsize
		
		\begin{center}
			\begin{tabular}{ l|c|c|c|c|c } 
				\toprule
				\textbf{System} &\textbf{Processed} &  \textbf{Recall} & \textbf{Precision} & \textbf{F-1} & \textbf{F-1 Global} \\
				\midrule
				\multirow{1}{*}{\textbf{HCqa}(our approach)} & $23$ & $\textbf{0.42}$&$0.42$&$\textbf{0.52}$&$\textbf{0.50}$\\
				\multirow{1}{*}{\textbf{Xu et.al. \cite{feng2016hybrid}}} & $-$ & $-$&$-$&$-$&$0.40$\\
				\multirow{1}{*}{\textbf{Xser}} & $23$ & $0.38$&$0.43$&$0.41$&$0.39$\\
				\multirow{1}{*}{\textbf{AskDBpedia}} & $21$ &$0.33$&$0.51$&$0.40$&$0.35$\\
				\multirow{1}{*}{\textbf{FirstRun LIMSI}} & $24$ & $0.04$&$0.61$&$0.08$&$0.08$\\
				
				\bottomrule
			\end{tabular}
		\end{center}
	\end{table}
	
	Next, we compare our system over the state-of-art systems listed in \cite{hu2018state} on the ComplexQuestion benchmark \cite{abujabal2017automated} using the average F1 metric. The average F1 of our system is higher than all the other systems except for STF which is customized for question answering on the knowledge graphs sources, while our approach is capable of answering questions over textual corpus and knowledge graphs.
	We also evaluated our system over the recently collected hybrid corpus\cite{grau2018corpus} which resulted 46.2\% for F-1 metric. 
	
	\begin{table}[hpt]\caption {The comparison study on the task of answer extraction on ComplexQuestion benchmark.} \label{tab:AnswerExtractionCQ}  
		
		\scriptsize
		\begin{center}
			\begin{tabular}{ l|c } 
				\toprule
				\textbf{System}	&\textbf{Average F-1} \\
				\midrule
				\multirow{1}{*}{\textbf{HCqa}  (our approach)} & $53.6$ \\\multirow{1}{*}{\textbf{STF}} & $\textbf{54.3}$\\
				\multirow{1}{*}{QUINT} & $49.2$\\
				\multirow{1}{*}{\textbf{Aqqu LIMSI}} & $27.8$ \\
				\multirow{1}{*}{\textbf{Aqqu++}} & $46.7$\\
				\bottomrule
			\end{tabular}
		\end{center}
	\end{table}

	\section{Conclusion and Future Work}
	In this paper, we presented  HCqa, a QA system dealing with complex questions and running on a hybrid of textual corpus and knowledge graphs.
	The proposed approach relies on extracting triple patterns and is able of working with both question and non-question sentences. It decomposes the given complex question into several sub-questions, which are triples. HCqa main contribution is presenting a novel and generic relation extraction approach for extracting sub-questions. We used a textual corpus\footnote{\url{http://www.mpi-inf.mpg.de/departments/d5/software/clausie}}, which was also employed by the prior art ClausIE \cite{DelCorro2013} and LS3RYIE \cite{vo2017self} to evaluate our relation extraction module. This corpus is compiled from three datasets (i) New York Times, (ii) Reverb and (iii) Wikipedia. In addition, we evaluated this module over inventory questions from the hybrid task of QALD-6 challenge\footnote{\url{https://qald.sebastianwalter.org/index.php?x=challenge&q=6}} \cite{unger20166th}. The module was compared for textual corpus as well as query inventory, with the recent work,  LS3RyIE  \cite{vo2017self}.
	The result showed some improvement for the relation extraction task over the baseline, using several defined settings. Besides we extracted other types of relation. Although more relations has been  extracted by our approach, the result showed more precision for all types of relations, rather than only for the baseline ones, over textual corpus as well as query inventory.\\ 
	Furthermore, HCqa represented two algorithms to federate answer set of sub-questions extracted in relation extraction module, from heterogeneous sources.

	We evaluated HCqa for the two essential tasks, (a) relation extraction, (b) answer set aggregation and (c) answer extraction.  Our experimental study exceeds the state-of-the-art as for the first task, it reaches the precision $81.74\%$ ($13.41\%$ improvement) on free text and $84.04\%$ ($39.36\%$ improvement) on QALD-6 query inventory.
	Also, the second task reaches the precision of $92.04\%$ on QALD-5 query inventory ($21.22\%$ improvement). These improvements result in higher F-score for answer extraction; $56\%$ ($30\%$ improvement) over QALD-5, and $50\%$ ($10\%$ improvement) over QALD-6 and shows comparable accuracy on ComplexQuestion benchmark.
	Our experiment shows that HCqa achieves the best performance among the hybrid QA systems.
	The implementation of HCqa is available at \url{https://github.com/asadifar/HCqa}. 
	We plan to extend this work using more number of knowledge graphs and to run it over more number of benchmarking datasets. Another plan is to implement this approach for domain-specific use cases such as the bio-medical domain.\\
	%\begingroup
	%\let\clearpage\relax
	
	%\nocite{*}
	
	    \bibliographystyle{plain}
	\bibliography{paper}

\begin{thebibliography}{10}

\bibitem{abacha2015means}
Asma~Ben Abacha and Pierre Zweigenbaum.
\newblock Means: A medical question-answering system combining nlp techniques
  and semantic web technologies.
\newblock {\em Information processing \& management}, 51(5):570--594, 2015.

\bibitem{abujabal2017automated}
Abdalghani Abujabal, Mohamed Yahya, Mirek Riedewald, and Gerhard Weikum.
\newblock Automated template generation for question answering over knowledge
  graphs.
\newblock In {\em Proceedings of the 26th international conference on world
  wide web}, pages 1191--1200. International World Wide Web Conferences
  Steering Committee, 2017.

\bibitem{bast2012broccoli}
Hannah Bast, Florian B{\"a}urle, Bj{\"o}rn Buchhold, and Elmar Haussmann.
\newblock Broccoli: Semantic full-text search at your fingertips.
\newblock {\em arXiv preprint arXiv:1207.2615}, 2012.

\bibitem{bast2007ester}
Holger Bast, Alexandru Chitea, Fabian Suchanek, and Ingmar Weber.
\newblock Ester: efficient search on text, entities, and relations.
\newblock In {\em Proceedings of the 30th annual international ACM SIGIR
  conference on Research and development in information retrieval}, pages
  671--678. ACM, 2007.

\bibitem{bhutani2016nested}
Nikita Bhutani, HV~Jagadish, and Dragomir Radev.
\newblock Nested propositions in open information extraction.
\newblock In {\em Proceedings of the 2016 Conference on Empirical Methods in
  Natural Language Processing}, pages 55--64, 2016.

\bibitem{bunescu2005shortest}
Razvan~C Bunescu and Raymond~J Mooney.
\newblock A shortest path dependency kernel for relation extraction.
\newblock In {\em Proceedings of the conference on human language technology
  and empirical methods in natural language processing}, pages 724--731.
  Association for Computational Linguistics, 2005.

\bibitem{cohen1960coefficient}
Jacob Cohen.
\newblock A coefficient of agreement for nominal scales.
\newblock {\em Educational and psychological measurement}, 20(1):37--46, 1960.

\bibitem{cunningham2002framework}
Hamish Cunningham, Diana Maynard, Kalina Bontcheva, and Valentin Tablan.
\newblock A framework and graphical development environment for robust nlp
  tools and applications.
\newblock In {\em ACL}, pages 168--175, 2002.

\bibitem{isem2013daiber}
Joachim Daiber, Max Jakob, Chris Hokamp, and Pablo~N Mendes.
\newblock Improving efficiency and accuracy in multilingual entity extraction.
\newblock In {\em Proceedings of the 9th International Conference on Semantic
  Systems}, pages 121--124. ACM, 2013.

\bibitem{decker2000}
Stefan Decker, Prasenjit Mitra, and Sergey Melnik.
\newblock Framework for the semantic web: an rdf tutorial.
\newblock {\em IEEE Internet Computing}, 4(6):68--73, 2000.

\bibitem{DelCorro2013}
Luciano Del~Corro and Rainer Gemulla.
\newblock Clausie: clause-based open information extraction.
\newblock In {\em Proceedings of the 22nd international conference on World
  Wide Web}, pages 355--366. ACM, 2013.

\bibitem{fader2011identifying}
Anthony Fader, Stephen Soderland, and Oren Etzioni.
\newblock Identifying relations for open information extraction.
\newblock In {\em Proceedings of the conference on empirical methods in natural
  language processing}, pages 1535--1545. Association for Computational
  Linguistics, 2011.

\bibitem{fader2013paraphrase}
Anthony Fader, Luke Zettlemoyer, and Oren Etzioni.
\newblock Paraphrase-driven learning for open question answering.
\newblock In {\em Proceedings of the 51st Annual Meeting of the Association for
  Computational Linguistics (Volume 1: Long Papers)}, volume~1, pages
  1608--1618, 2013.

\bibitem{PaoloFerragina2010}
Paolo Ferragina and Ugo Scaiella.
\newblock Tagme: on-the-fly annotation of short text fragments (by wikipedia
  entities).
\newblock In {\em Proceedings of the 19th ACM international conference on
  Information and knowledge management}, pages 1625--1628. ACM, 2010.

\bibitem{frost2014event}
Richard~A Frost, Wale Agboola, Eric Matthews, and Jonathan~A Donais.
\newblock An event-driven approach for querying graph-structured data using
  natural language.
\newblock In {\em EDBT/ICDT Workshops}, volume 2014, pages 192--199, 2014.

\bibitem{frost2014denotational}
Richard~A Frost, Jonathon Donais, E~Matthews, and Rob Stewart.
\newblock A denotational semantics for natural langauge query interfaces to
  semantic web triplestores.
\newblock {\em Submitted for publication}, pages 18--20, 2014.

\bibitem{grau2018corpus}
Brigitte Grau and Anne-Laure Ligozat.
\newblock A corpus for hybrid question answering systems.
\newblock In {\em Companion of the The Web Conference 2018 on The Web
  Conference 2018}, pages 1081--1086. International World Wide Web Conferences
  Steering Committee, 2018.

\bibitem{harabagiu2005employing}
Sanda~M Harabagiu, Dan~I Moldovan, Christine Clark, Mitchell Bowden, Andrew
  Hickl, and Patrick Wang.
\newblock Employing two question answering systems in trec 2005.
\newblock In {\em TREC}, 2005.

\bibitem{hickl2004experiments}
Andrew Hickl, John Lehmann, John Williams, and Sanda Harabagiu.
\newblock Experiments with interactive question answering in complex scenarios.
\newblock In {\em Proceedings of the Workshop on Pragmatics of Question
  Answering at HLT-NAACL 2004}, 2004.

\bibitem{hu2018state}
Sen Hu, Lei Zou, and Xinbo Zhang.
\newblock A state-transition framework to answer complex questions over
  knowledge base.
\newblock In {\em Proceedings of the 2018 Conference on Empirical Methods in
  Natural Language Processing}, pages 2098--2108, 2018.

\bibitem{huddleston2002cambridge}
Rodney Huddleston and Geoffrey~K Pullum.
\newblock The cambridge grammar of english.
\newblock {\em Language. Cambridge: Cambridge University Press}, pages 1--23,
  2002.

\bibitem{jespersen2003essentials}
Otto Jespersen.
\newblock {\em Essentials of English Grammar: 25th impression, 1987}.
\newblock Routledge, 2003.

\bibitem{Kolomiyets2011}
Oleksandr Kolomiyets and Marie-Francine Moens.
\newblock A survey on question answering technology from an information
  retrieval perspective.
\newblock {\em Information Sciences}, 181(24):5412--5434, 2011.

\bibitem{lopez2009merging}
Vanessa Lopez, Andriy Nikolov, Miriam Fernandez, Marta Sabou, Victoria Uren,
  and Enrico Motta.
\newblock Merging and ranking answers in the semantic web: The wisdom of
  crowds.
\newblock In {\em Asian Semantic Web Conference}, pages 135--152. Springer,
  2009.

\bibitem{lopez2009cross}
Vanessa Lopez, Victoria Uren, Marta~Reka Sabou, and Enrico Motta.
\newblock Cross ontology query answering on the semantic web: an initial
  evaluation.
\newblock In {\em Proceedings of the fifth international conference on
  Knowledge capture}, pages 17--24. ACM, 2009.

\bibitem{miller2013}
Brad Miller and David Ranum.
\newblock Problem solving with algorithms and data structures, 2013.

\bibitem{kilgarriff2000wordnet}
George Miller.
\newblock {\em WordNet: An electronic lexical database}.
\newblock MIT press, 1998.

\bibitem{oh2011compositional}
Hyo-Jung Oh, Ki-Youn Sung, Myung-Gil Jang, and Sung~Hyon Myaeng.
\newblock Compositional question answering: A divide and conquer approach.
\newblock {\em Information Processing \& Management}, 47(6):808--824, 2011.

\bibitem{park2015isoft}
Seonyeong Park, Soonchoul Kwon, Byungsoo Kim, and Gary~Geunbae Lee.
\newblock Isoft at qald-5: Hybrid question answering system over linked data
  and text data.
\newblock In {\em CLEF (Working Notes)}, 2015.

\bibitem{saquete2004splitting}
Estela Saquete, Patricio Martinez-Barco, Rafael Munoz, and Jose-Luis Vicedo.
\newblock Splitting complex temporal questions for question answering systems.
\newblock In {\em Proceedings of the 42nd Annual Meeting on Association for
  Computational Linguistics}, page 566. Association for Computational
  Linguistics, 2004.

\bibitem{schmitz2012open}
Michael Schmitz, Robert Bart, Stephen Soderland, Oren Etzioni, et~al.
\newblock Open language learning for information extraction.
\newblock In {\em Proceedings of the 2012 Joint Conference on Empirical Methods
  in Natural Language Processing and Computational Natural Language Learning},
  pages 523--534. Association for Computational Linguistics, 2012.

\bibitem{Schuster2016}
Sebastian Schuster and Christopher~D Manning.
\newblock Enhanced english universal dependencies: An improved representation
  for natural language understanding tasks.
\newblock In {\em LREC}, pages 23--28. Portoro{\v{z}}, Slovenia, 2016.

\bibitem{sina2}
Saeedeh Shekarpour, Edgard Marx, Axel-Cyrille~Ngonga Ngomo, and S{\"o}ren Auer.
\newblock Sina: Semantic interpretation of user queries for question answering
  on interlinked data.
\newblock {\em Web Semantics: Science, Services and Agents on the World Wide
  Web}, 30:39--51, 2015.

\bibitem{sina1}
Saeedeh Shekarpour, Axel-Cyrille Ngonga~Ngomo, and S{\"o}ren Auer.
\newblock Question answering on interlinked data.
\newblock In {\em Proceedings of the 22nd international conference on World
  Wide Web}, pages 1145--1156. ACM, 2013.

\bibitem{singh2018frankenstein}
Kuldeep Singh, Andreas Both, Arun Sethupat, and Saeedeh Shekarpour.
\newblock Frankenstein: A platform enabling reuse of question answering
  components.
\newblock In {\em European Semantic Web Conference}, pages 624--638. Springer,
  2018.

\bibitem{Frankenstein2}
Kuldeep Singh, Andreas Both, Arun Sethupat, and Saeedeh Shekarpour.
\newblock Frankenstein: A platform enabling reuse of question answering
  components.
\newblock In {\em European Semantic Web Conference}, pages 624--638. Springer,
  2018.

\bibitem{Frankenstein1}
Kuldeep Singh, Arun~Sethupat Radhakrishna, Andreas Both, Saeedeh Shekarpour,
  Ioanna Lytra, Ricardo Usbeck, Akhilesh Vyas, Akmal Khikmatullaev, Dharmen
  Punjani, Christoph Lange, et~al.
\newblock Why reinvent the wheel: Let's build question answering systems
  together.
\newblock In {\em Proceedings of the 2018 World Wide Web Conference on World
  Wide Web}, pages 1247--1256. International World Wide Web Conferences
  Steering Committee, 2018.

\bibitem{Toutanova2013}
Kristina Toutanova and Christopher~D Manning.
\newblock Enriching the knowledge sources used in a maximum entropy
  part-of-speech tagger.
\newblock In {\em Proceedings of the 2000 Joint SIGDAT conference on Empirical
  methods in natural language processing and very large corpora: held in
  conjunction with the 38th Annual Meeting of the Association for Computational
  Linguistics-Volume 13}, pages 63--70. Association for Computational
  Linguistics, 2000.

\bibitem{unger20155th}
Christina Unger, Corina Forascu, Vanessa Lopez, Axel-Cyrille Ngomo, Elena
  Cabrio, Philippand Cimiano, and Sebastian Walter.
\newblock Question answering over linked data (qald-5).
\newblock In {\em Question Answering over Linked Data (QALD-5)}. CLEF Working
  Notes, 2015.

\bibitem{unger2014question}
Christina Unger, Corina Forascu, Vanessa Lopez, Axel-Cyrille~Ngonga Ngomo,
  Elena Cabrio, Philipp Cimiano, and Sebastian Walter.
\newblock Question answering over linked data (qald-4).
\newblock In {\em Working Notes for CLEF 2014 Conference}, 2014.

\bibitem{unger20166th}
Christina Unger, Axel-Cyrille~Ngonga Ngomo, and Elena Cabrio.
\newblock 6th open challenge on question answering over linked data (qald-6).
\newblock In {\em Semantic Web Evaluation Challenge}, pages 171--177. Springer,
  2016.

\bibitem{usbeck2015hawk}
Ricardo Usbeck, Axel-Cyrille~Ngonga Ngomo, Lorenz B{\"u}hmann, and Christina
  Unger.
\newblock Hawk--hybrid question answering using linked data.
\newblock In {\em European Semantic Web Conference}, pages 353--368. Springer,
  2015.

\bibitem{vo2017self}
Duc-Thuan Vo and Ebrahim Bagheri.
\newblock Self-training on refined clause patterns for relation extraction.
\newblock {\em Information Processing \& Management}, 54(4):686--706, 2018.

\bibitem{wu2010open}
Fei Wu and Daniel~S Weld.
\newblock Open information extraction using wikipedia.
\newblock In {\em Proceedings of the 48th annual meeting of the association for
  computational linguistics}, pages 118--127. Association for Computational
  Linguistics, 2010.

\bibitem{feng2016hybrid}
Kun Xu, Yansong Feng, Songfang Huang, and Dongyan Zhao.
\newblock Hybrid question answering over knowledge base and free text.
\newblock In {\em Proceedings of COLING 2016, the 26th International Conference
  on Computational Linguistics: Technical Papers}, pages 2397--2407, 2016.

\bibitem{yates2007textrunner}
Alexander Yates, Michael Cafarella, Michele Banko, Oren Etzioni, Matthew
  Broadhead, and Stephen Soderland.
\newblock Textrunner: open information extraction on the web.
\newblock In {\em Proceedings of Human Language Technologies: The Annual
  Conference of the North American Chapter of the Association for Computational
  Linguistics: Demonstrations}, pages 25--26. Association for Computational
  Linguistics, 2007.

\bibitem{zandvoort2001handbook}
Reinard~Willem Zandvoort.
\newblock {\em A handbook of English grammar}.
\newblock Longman, 2001.

\end{thebibliography}
	
	%\endgroup
\end{document}